\documentclass[11pt]{article}

    \PassOptionsToPackage{table}{xcolor}

\usepackage[preprint]{acl}
\usepackage{predact-lineno-compat}
\usepackage{times}
\usepackage{latexsym}
\usepackage{multirow}
\usepackage{booktabs}
\usepackage{adjustbox}
\usepackage{pifont}
\usepackage{tikz}
\usepackage{xspace}
\usepackage{listings}
\usepackage{amsmath}
\usepackage{multirow}
\usepackage{fvextra}
\usepackage{dsfont}
\usepackage[breakable,skins]{tcolorbox}
\usepackage{wrapfig}
\usepackage[table]{xcolor}
\newcommand{\cmark}{\textcolor{green!60!black}{\ding{51}}}
\newcommand{\xmark}{\textcolor{red}{\ding{55}}}

\newcommand{\pmark}{\textcolor{orange!80!black}{\ding{108}}} 

\usepackage{booktabs}
\usepackage{makecell}
\usepackage{xcolor}
\usepackage{pifont}
\usepackage{graphicx}
\usepackage{svg}
\usepackage{booktabs}
\usepackage[utf8]{inputenc} 
\usepackage[T1]{fontenc}    
\usepackage{hyperref}       
\usepackage{etoc}           
\usepackage{xurl}           
\usepackage{booktabs}       
\usepackage{amsfonts}       
\usepackage{nicefrac}       
\usepackage{microtype}      
\usepackage{xcolor}         

\newcommand{\predact}{\textsc{PredActBench}\xspace}
\newcommand{\predactcs}{\textsc{PredAct-CS}\xspace}

\title{\textsc{PredActBench}: Benchmarking Tool-Augmented Dialogue under Controlled Tool Noise
}

\author{%
  Abdulrahman AlRabah\thanks{Corresponding author.} \quad
  Xiaocheng Yang \quad
  Dilek Hakkani-T\"{u}r \quad
  Abdussalam Alawini \\
  University of Illinois Urbana-Champaign \\
  \texttt{\{alrabah2, xy61, dilek, alawini\}@illinois.edu}
}

\newcommand{\answerTODO}[1][]{\textcolor{red}{\bf [TODO]}}
\newcommand{\justificationTODO}[1][]{\textcolor{red}{\bf [TODO]}}

\begin{document}
\etocdepthtag.toc{mainmatter}
\raggedbottom

\maketitle

\vspace{-1mm}
\begin{abstract}
Large language model (LLM) agents are increasingly used to support decisions by calling external tools. Existing work on LLM tool-calling assumes these tools are perfectly reliable. However, in many applications, the tools  are statistical predictors with non-trivial errors. We introduce \predact, a benchmark for evaluating dialogue agents that work with imperfect predictive tools, using education as a testbed. First, we build a benchmark in which LLMs work with noisy predictors to guide users' decisions. Second, we extend two prior metrics for trust calibration, Relative AI-Reliance and Relative Self-Reliance, to multi-turn dialogue. Third, we evaluate 13 LLMs on OULAD and our released \predactcs dataset of 60 courses with real final grades and synthetic weekly scores, alongside a human study involving 13 instructors and teaching assistants. We find that when tools are noisy, SOTA models rarely make predictor uncertainty visible to users, with our strongest baseline \mbox{GPT-5.5} achieving an F1 of 57.2.  As a result, users over-rely on incorrect predictions and unsupported claims, showing that task accuracy alone is not enough for AI decision support and agents must communicate when their tools may be wrong. We release \predact\footnote{\begin{tabular}[t]{@{}l@{\ }l@{}}Code: & \href{https://github.com/aalrabah/PredAct_bench}{github.com/aalrabah/PredAct\_bench}\\ Data: & \href{https://huggingface.co/datasets/predact/predact-cs}{huggingface.co/datasets/predact/predact-cs}\end{tabular}} to support the development of LLM agents that help users calibrate their reliance.
\end{abstract}
\vspace{-1mm}

\section{Introduction}

\begin{figure}[!t]
  \centering
  \includegraphics[width=0.94\columnwidth]{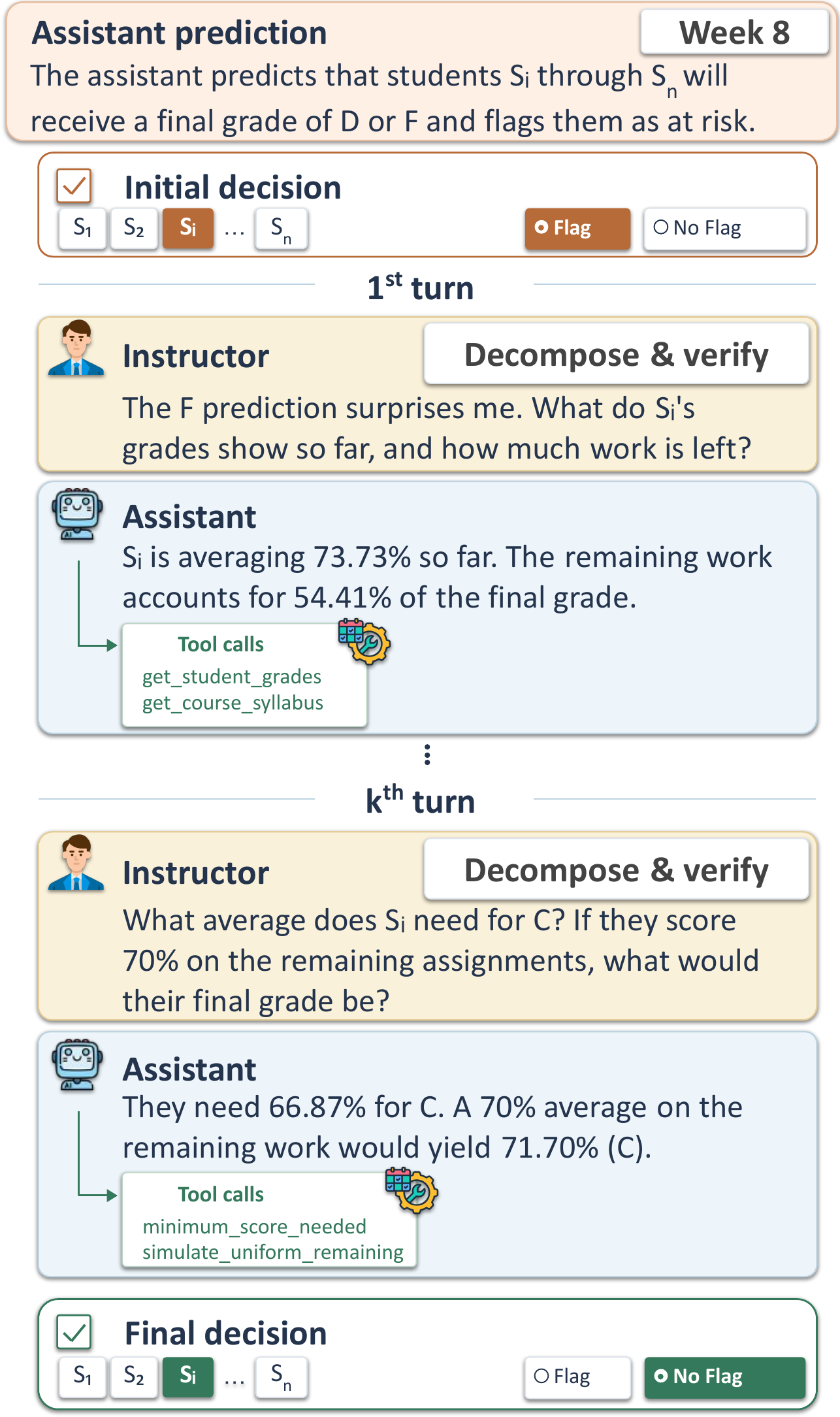}
  \caption{An illustrative \predact episode showing initial decisions, $k$ dialogue turns backed by tool calls, and final decisions.}
  \label{fig:intro_dialogue}
\end{figure}

As grades accumulate over a semester, instructors must continually decide which students need help and what kind of help they need. This is especially challenging in large courses; instructors must track who is struggling, decide who needs feedback, and work out what each student still needs to do to succeed. LLM assistants with access to students' grades and predictive models could ease this burden, but only if instructors can tell what the assistant knows from what it is guessing.
For example, in the dialogue of Figure~\ref{fig:intro_dialogue}, the instructor checks a student’s grades and remaining work to see what scores are needed from week 8 onward to pass despite a predicted F. The instructor builds their verification judgment by asking more questions to decide when and how to intervene. The underlying challenge is broader than education. In any domain with evolving evidence, we need to distinguish reliable data from uncertain predictions and find a way to share only reliable information with humans for better judgment.

Existing dialogue benchmarks mainly evaluate state tracking and response generation under static user goals~\cite{budzianowski-etal-2018-multiwoz, eric-etal-2020-multiwoz, zang-etal-2020-multiwoz, 10.1007/978-3-030-88483-3_16, ye-etal-2022-multiwoz, Rastogi_Zang_Sunkara_Gupta_Khaitan_2020}. Tool-use and agent benchmarks emphasize action correctness, policy adherence, or environment interaction~\cite{qin2024toolllm, NEURIPS2024_e4c61f57, patil2025the, yao2025taubench, zhou2023webarena, deng2023mindweb}. Robustness benchmarks study noisy instructions and imperfect tool execution~\cite{kim2026perfectapiscomprehensiveevaluation, wang2026agentnoisebenchbenchmarkingrobustnesstoolusing}. However, these lines of work largely treat these three challenges in isolation, although they co-occur in real deployments:  reasoning over partial longitudinal evidence, maintaining coherent state through multi-turn conversation, and making actionable decisions when tool outputs are uncertain.

We introduce \predact, an education-focused benchmark for evaluating predictive, interactive, and tool-augmented dialogue agents. In each episode, given partial student trajectories up to a temporal cutoff, an agent must predict final outcomes, identify at-risk students, and recommend interventions through multi-turn interactions with an instructor or teaching assistant. \predact includes both system-initiated analysis and user-initiated dialogue, where instructors query, filter, and challenge the agent's analysis. To reflect realistic decision support, \predact combines deterministic lookup and counterfactual tools with probabilistic prediction tools whose reliability varies across courses and semester stages.
Our experiments analyze whether current LLM agents can remain reliable when temporal reasoning, noisy tools, and human decision-making are coupled. We evaluate grade prediction accuracy, decision quality, tool calibration, and human trust calibration.

Our contributions are threefold:
\begin{itemize}
    \item We introduce \predact, a benchmark for evaluating dialogue agents under controlled tool noise, and release \predactcs, a dataset covering 60 computer science (CS) courses with real final grades and synthetically generated weekly scores. We evaluate \predact on \predactcs and OULAD \citep{kuzilek2017}, which provides real assessment trajectories.
    \vspace{-1mm}

    \item We evaluate decision quality using initial and final F1 scores, and introduce episode-level RAIR and RSR to extend the trust-calibration framework of \citet{schemmer2023appropriate} from one-shot decisions to multi-turn dialogue.
    \vspace{-1mm}

    \item We evaluate 13 closed and open source LLMs across both datasets, alongside a human study with 13 instructors and teaching assistants. LLM agents' F1 generally improves with tool accuracy, while RAIR/RSR reveal better-calibrated human reliance.
\end{itemize}

\section{Related Work}


\begin{table*}[t]
\centering
\small
\renewcommand{\cmark}{\ding{51}}
\renewcommand{\xmark}{\ding{55}}
\renewcommand{\pmark}{\ding{108}}
\setlength{\tabcolsep}{3pt}
\renewcommand{\arraystretch}{1.0}
\begin{tabular*}{\textwidth}{@{\extracolsep{\fill}}l>{\columncolor{gray!10}}cccccc@{}}
\toprule
\textbf{Feature} & \textbf{\predact} & \makecell{\textbf{MultiWOZ}\\\textbf{2.4}} & \makecell{\textbf{TIMER}\\\textbf{Bench}} & \makecell{\textbf{$\tau^2$}\\\textbf{bench}} & \makecell{\textbf{AgentNoise}\\\textbf{Bench}} & \makecell{\textbf{WildAGT}\\\textbf{Eval}} \\
\midrule
Tool Calls & \cmark & \xmark & \xmark & \cmark & \cmark & \cmark \\
Multi-Turn Conversation & \cmark & \cmark & \xmark & \cmark & \cmark & \cmark \\
Interactive Conversation & \cmark & \xmark & \xmark & \cmark & \cmark & \cmark \\
Noise Injection & \cmark & \xmark & \xmark & \xmark & \cmark & \cmark \\
Longitudinal Reasoning & \cmark & \xmark & \cmark & \xmark & \xmark & \xmark \\
Human-in-the-Loop & \cmark & \xmark & \pmark & \xmark & \xmark & \xmark \\
\midrule
Task Domain & Education & \makecell{Attraction,\\Hospital,\\Etc.} & \makecell{Clinical\\EHR} & \makecell{Airline,\\Retail,\\Telecom} & \makecell{Airline,\\Retail,\\Telecom} & \makecell{Time Notification,\\Communication,\\Etc.} \\
\bottomrule
\end{tabular*}
\caption{Benchmark comparison. \cmark{} present, \xmark{} absent, and \pmark{} expert validation without live human interaction. Benchmark references are \cite{ye-etal-2022-multiwoz,cui2025timer,barres2025tau2benchevaluatingconversationalagents,wang2026agentnoisebenchbenchmarkingrobustnesstoolusing,kim2026perfectapiscomprehensiveevaluation}.}
\label{tab:benchmark_comparison}
\end{table*}

\paragraph{Dialogue and Agent Benchmarks}

MultiWOZ~\cite{budzianowski-etal-2018-multiwoz, eric-etal-2020-multiwoz, zang-etal-2020-multiwoz, 10.1007/978-3-030-88483-3_16, ye-etal-2022-multiwoz} and Schema-Guided Dialogue (SGD)~\cite{Rastogi_Zang_Sunkara_Gupta_Khaitan_2020} standardize belief state tracking and response generation under predefined schemas and static goals. APIBench~\cite{NEURIPS2024_e4c61f57}, ToolBench~\cite{qin2024toolllm}, and BFCL~\cite{patil2025the} evaluate API selection, argument construction, compositional planning, and multi-turn function calling, while WebArena~\cite{zhou2023webarena}, MIND2WEB~\cite{deng2023mindweb}, $\tau$-bench~\cite{yao2025taubench}, and PrefIx~\cite{li2026prefixunderstandadaptuser} extend evaluation to web tasks, simulated user--agent--API interaction, policy adherence, and user preference adaptation. WILDAGTEVAL~\cite{kim2026perfectapiscomprehensiveevaluation} and AgentNoiseBench~\cite{wang2026agentnoisebenchbenchmarkingrobustnesstoolusing} evaluate agents under realistic API complexity and controllable user- or tool-side noise. These lines of work do not jointly evaluate longitudinal prediction, belief tracking over evolving evidence, noisy probabilistic tools, and human decision support in high-stakes educational settings.

\paragraph{Robust Agents under Noise}

Prior work studies ambiguity clarification~\citep{zhang-etal-2024-clamber}, incomplete tool-use conditions~\citep{yang2025toolaugmentedlargelanguagemodels}, tool hallucination and reliability failures~\citep{zhang-etal-2024-toolbehonest, xu2025reducingtoolhallucinationreliability}, parameter-filling and toolchain errors~\citep{xiong-etal-2025-butterfly}, and adversarial tool injection~\citep{zhang-etal-2025-allies}. WILDAGTEVAL~\citep{kim2026perfectapiscomprehensiveevaluation} and AgentNoiseBench~\citep{wang2026agentnoisebenchbenchmarkingrobustnesstoolusing} expose agents to execution failures, incomplete responses, erroneous outputs, misleading signals, and irrelevant information, while PALADIN studies recovery from execution-level tool failures~\citep{vuddanti2025paladinselfcorrectinglanguagemodel}. Agents must recognize solvability, ground parameters reliably, recover from failures, and remain stable under both benign and adversarial perturbations.

\paragraph{LLM Temporal Reasoning}

LLMs remain limited in temporal reasoning~\cite{su2024timo, chu2024timebench}. TimeBench~\cite{chu2024timebench}, TIMER-Bench~\cite{cui2025timer}, and ER-Reason~\cite{mehandru2025erreasonbenchmarkdatasetllmbased} reveal gaps between models and humans across symbolic, commonsense, clinical, and event-level temporal reasoning. Prior work improves temporal reasoning through prompting, consistency-based methods, parameter adaptation, temporal or mathematical training signals~\cite{su2024timo, kim-hwang-2025-counterfactual, nylund-etal-2024-time}, and structure-aware integration of temporal knowledge graphs with text representations~\cite{jiang-etal-2025-towards-explainable}. Models exhibit chronological confusion, recency bias, and failures to integrate information across time steps~\cite{cui2025timer, porter2024llmd}. Therefore, our work evaluates how agents use dialogue to query partial student data and verify predictions with noisy tools. 

\paragraph{Evaluation Metrics for Probabilistic and Human-AI Decision Support}

Brier score measures the squared error of probabilistic forecasts~\citep{glenn1950verification}, while Expected Calibration Error (ECE) summarizes the gap between predicted confidence and empirical accuracy~\citep{10.5555/3305381.3305518}. Appropriate reliance on AI advice requires users to follow correct recommendations while rejecting incorrect ones~\citep{schemmer2023appropriate}. In this work, we assess our predictor's calibration using ECE and Brier score and extend RAIR and RSR to a multi-turn dialogue evaluation.

\section{\predact}

\noindent\textbf{Task Definition. } \predact formulates academic risk
prediction as a dialogue task over multiple turns between an
\emph{LLM instructor}, which simulates a course instructor, and an
\emph{LLM Assistant}, which responds through tool calls. Both are
released as part of the benchmark. Each episode is parameterized by a
course, a cutoff week $t$ (only grades recorded up to week $t$ are used for prediction), and a sample of $N$ students. Given a partial grade trajectory $\mathbf{x}_i^{(t)}$ for each student $i$, the assistant predicts a final letter grade $\hat{y}_i$ with confidence $c_i$, flags students it believes are at risk, and surfaces supporting
evidence. The interaction proceeds in three stages. In the first stage, the assistant presents its flagged students with confidence scores and brief rationales. In the second stage, the instructor records an initial decision $d_i^{\text{init}} \in \{\text{flag}, \text{no\_flag}\}$ based solely on this report. In the third stage,
the instructor engages in further dialogue with the assistant through a chat interface shared by human and LLM instructors, probing grade history, class comparisons, counterfactuals, and interventions, before committing to a final decision $d_i^{\text{final}}$. The instructor retains final authority, and the assistant acts as a collaborator augmented by tools rather than an autonomous agent.

\begin{figure*}[t]

  \centering
  \includegraphics[width=\textwidth]{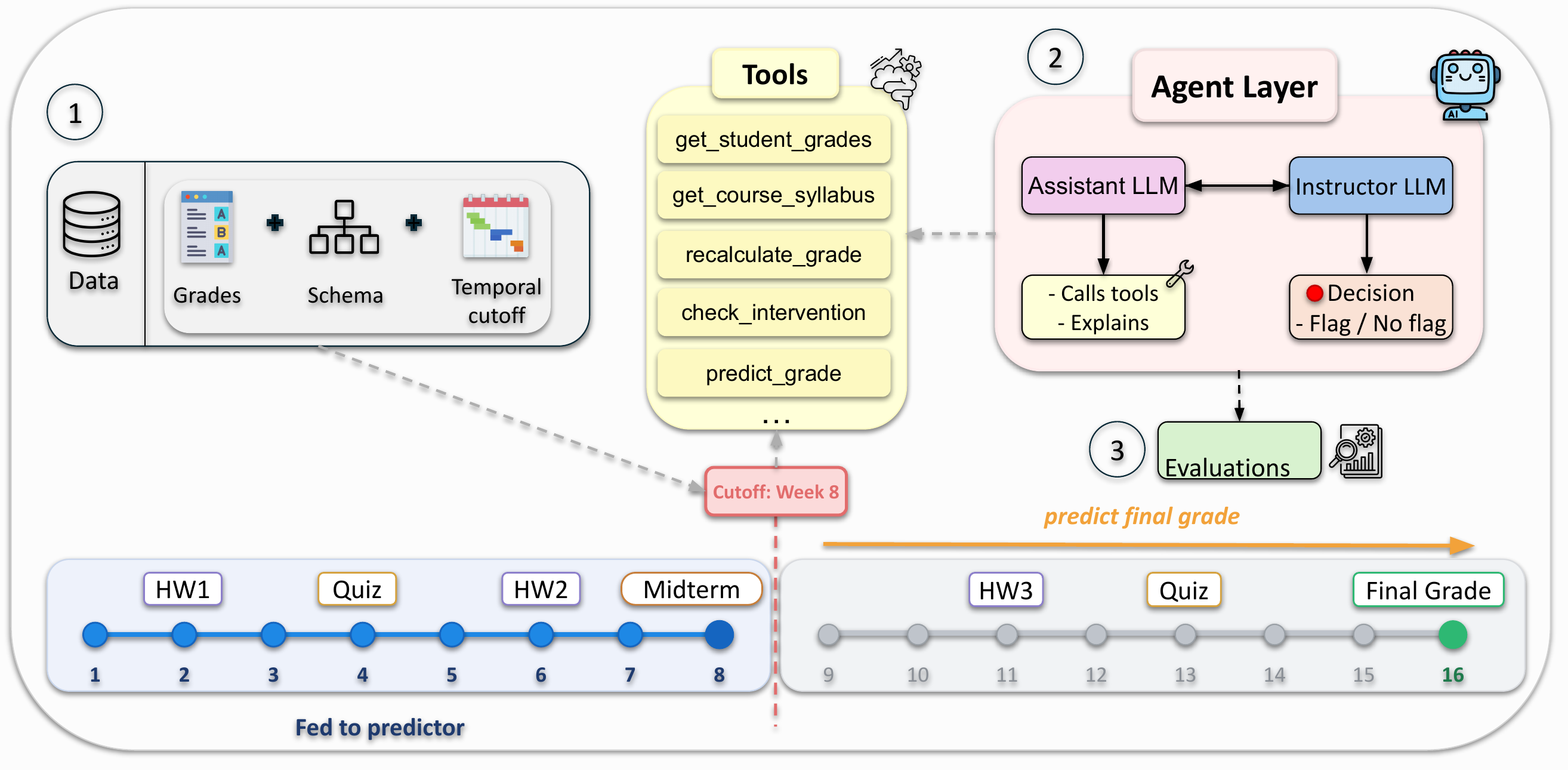}
  \caption{System-initiated dialogue between two LLM agents (left) and user-initiated dialogue with an instructor (right) in \predact. Both use tools to analyze student performance and intervention decisions. Red text shows tool outputs.}
  \label{fig:arch}
\end{figure*}

\subsection{Benchmark Overview}
In this section, we show what goes in and out of \predact, as shown in Figure~\ref{fig:arch}. First, we collect data containing student grades and course schema, and choose the week at which to set the temporal cutoff. Second, an assistant LLM and an instructor (LLM or human) discuss the predictor's output, with the assistant querying \textit{deterministic} and \textit{probabilistic} tools described in Appendix~\ref{app:repro:tools} to help verify the predictions. Temporal reasoning here means asking questions and verifying output trends using these tools. Third, decisions are evaluated using F1, precision, recall, and the trust calibration metrics RAIR and RSR.

\subsection{Datasets}

\begin{table}[!htbp]
\centering
\small
\setlength{\tabcolsep}{4pt}
\renewcommand{\arraystretch}{1.0}
\caption{Dataset statistics for OULAD and \predactcs.}
\label{tab:datasets}
\begin{tabular*}{\columnwidth}{@{\extracolsep{\fill}}lll@{}}
\toprule
\textbf{Statistic} & \textbf{\predactcs} & \textbf{OULAD} \\
\midrule
Students & 53,401 & 32,593 \\
Courses & 60 CS courses & \makecell[l]{7 modules,\\22 presentations} \\
Final Grades & Real & Real \\
Trajectories & Synthetic (calibrated) & Real \\
\bottomrule
\end{tabular*}
\end{table}


The \predactcs dataset is derived from 60 undergraduate and graduate computer science courses at a large public research university. All assessment activities are captured with weights and due dates, paired with each student's real final grade (A to F scale). Student final grades are publicly available, yet raw weekly trajectories are not, so we synthetically create each student's trajectory to map to their final grade. OULAD \citep{kuzilek2017} also provides real assessment scores paired with each student's final grade (A to F scale). We split both datasets 80/20 using a fixed random seed, with training data serving as a historical record pool for the kNN prediction tool. We release our dataset and provide all details in Appendix~\ref{app:repro}.

\noindent\textbf{Tools. }\predact provides the agent with 12 tools across four categories, as shown in Table~\ref{tab:tools} in Appendix~\ref{app:repro:tools}. Prior benchmarks such as $\tau$-bench \citep{yao2025taubench} use tools as oracles, while AgentNoiseBench \citep{wang2026agentnoisebenchbenchmarkingrobustnesstoolusing} explores noise in user queries and tool outputs. In \predact, noise is placed in the toolchain, which challenges the agent and its human collaborator to reason when tool outputs are not quite accurate, with all tool details listed in Appendix~\ref{app:repro:tools}.

\begin{table*}[t]
\centering
\footnotesize
\setlength{\tabcolsep}{12pt}
\renewcommand{\arraystretch}{1.1}
\caption{Per-episode behavior profile on \predactcs. \textbf{Turns}: mean messages. \textbf{Duration}: mean seconds. \textbf{Q-types}: mean distinct question categories used (of 6). \textbf{Override \%}: agent-flagged students whose initial decision changed after chat. Humans: 117 episodes; LLMs: 50 episodes each.}
\label{tab:behavior_profile}
\begin{tabular*}{\textwidth}{@{\extracolsep{\fill}} l cccc @{}}
\toprule
\textbf{Rater} & \textbf{Turns} & \textbf{Duration (s)} & \textbf{Q-types (of 6)} & \textbf{Override \%} \\
\midrule
\rowcolor{gray!20} \multicolumn{5}{c}{\textit{Closed-source LLMs}} \\
\midrule
GPT-5.5               & \textbf{5.04}  & 185.2 & \textbf{4.02} & \textbf{15.1} \\
GPT-5.4 Mini          & 9.86  & 104.3 & 3.76 & 10.7 \\
GPT-4o Mini           & 10.00 & 86.4  & 3.00 & 19.2 \\
Claude Opus 4.7       & 8.96  & 135.4 & 3.12 & 18.5 \\
Claude Haiku 4.5      & 9.26  & 186.6 & 3.80 & 27.4 \\
Gemini 3.1 Pro        & 4.84  & 168.6 & 2.30 & 19.6 \\
Gemini 3 Flash        & 7.78  & 103.3 & 3.20 & 16.7 \\
\midrule
\rowcolor{gray!20} \multicolumn{5}{c}{\textit{Open-source LLMs}} \\
\midrule
DeepSeek V4 Pro       & 8.48  & 321.7 & 3.24 & 18.1 \\
DeepSeek V4 Flash     & 9.20  & 179.9 & 3.74 & 20.8 \\
Mistral Small 24B     & 9.56  & 145.0 & 3.76 & 14.9 \\
Ministral 3 14B       & 9.50  & 126.6 & 4.40 & 26.2 \\
Qwen 3.5 35B          & 8.84  & 230.3 & 3.46 & 13.5 \\
Qwen 3.5 9B           & 9.64  & 110.2 & 3.28 & 25.7 \\
\midrule
\rowcolor{gray!20} \multicolumn{5}{c}{\textit{Human participants ($n=13$)}} \\
\midrule
Humans (no-agent baseline) & --- & 175.0 & --- & --- \\
Humans (with agent)        & 3.65 & 302.1 & 2.49 & 20.7 \\
\bottomrule
\end{tabular*}
\end{table*}

\subsection{Grade Prediction and Noise Injection}
\label{sec:knn_noise}

\noindent\textbf{k-Nearest Neighbors Predictor.} Prior work predicts final course grades from partial assessment scores using historical student records~\cite{meier2015predicting}. The prediction tools used by the agent are designed as a weighted $k$-NN over historical student records, so we keep the prediction method fixed to evaluate how instructors use its advice. Given a student trajectory $\mathbf{x}_i^{(t)}$ at cutoff week $t$, we extract graded scores for each component observed in weeks $\leq t$. For each historical student $j$ in the training pool, we compute a Manhattan style averaged score distance over shared components.
\begin{equation}
d(i, j) = \frac{1}{|S_{ij}|} \sum_{p \in S_{ij}} | x_{i,p}^{(t)} - x_{j,p} |
\end{equation}
where $S_{ij}$ is the set of components both students share. We apply a radius search and retain all 
historical students $j$ with $d(i, j) \leq \tau_{ij}$, where the adaptive tolerance is
\begin{equation}
\tau_{ij} = \mathrm{clip}(0.5 \cdot \bar{\sigma}_S, 3, 10) \cdot 
            (0.4 + 0.6 \cdot r_{ij})
\end{equation}
with $\bar{\sigma}_S$ the mean per component standard deviation 
across the training pool and $r_{ij} = |S_{ij}| / |\text{components}|$ 
the share of components covered by both students. The predicted grade 
$\hat{y}_i$ is computed by distance weighted voting,
\begin{equation}
\begin{gathered}
\hat{y}_i = \arg\max_{g} \sum_{j \in \mathcal{N}_i} w_{ij} \cdot 
            \mathds{1}[y_j = g], \\
\quad w_{ij} = \frac{1}{1 + d(i, j)}
\end{gathered}
\end{equation}
where $\mathcal{N}_i$ is the set of matched neighbors and $y_j$ is 
neighbor $j$'s known final grade. The confidence is the share of 
weighted votes assigned to the predicted grade.
\begin{equation}
c_i = \frac{\sum_{j \in \mathcal{N}_i} w_{ij} \cdot 
            \mathds{1}[y_j = \hat{y}_i]}{\sum_{j \in \mathcal{N}_i} w_{ij}}.
\end{equation}
When no historical students fall within the tolerance, the predictor 
falls back to a score conditional grade distribution estimated from 
the training pool. In our experiments, the fallback fires for 19\% of 
OULAD predictions and 0\% of \predactcs predictions.
OULAD trajectories are sparse and irregular, whereas \predactcs contains complete synthetic assignment records. Both datasets use the same accuracy calibration, but the synthetic trajectories remain more regular than the real ones.

\noindent\textbf{Noise Injection.} To evaluate dialogue agents under 
controlled tool unreliability, \predact introduces calibrated noise 
into the predictor outputs. Given a sample of $N$ predictions with 
$n_0$ naturally correct, we calibrate to a target accuracy 
$a^* \in \{0.4, 0.5, 0.6, 0.7, 0.8\}$ by flipping a fixed number of 
predictions:
\begin{equation}
n_{\text{flip}} = | \mathrm{round}(a^* \cdot N) - n_0 |
\end{equation}
If $n_0 > a^* \cdot N$, we shuffle the correct predictions and flip 
the first $n_{\text{flip}}$ to a uniformly random wrong grade chosen 
from the four remaining letter grades. If $n_0 < a^* \cdot N$, we 
shuffle the wrong predictions and flip the first $n_{\text{flip}}$ 
to the truth label. Only $\hat{y}_i$ is rewritten; $c_i$ is 
preserved, so confidence and predicted grade can be inconsistent 
after injection (a deliberate property mirroring the miscalibration 
of real predictors). Tool side functions (grade lookups, 
counterfactuals, minimum score queries) are unaffected and always 
return ground truth derived values, isolating noise to the 
prediction tool only.
Uniform flipping controls accuracy without assuming that real errors are uniformly distributed. Section~\ref{sec:ablations} tests errors concentrated near the C grade boundary.

\section{Experimental Setup}


\noindent\textbf{Implementation Details.} We fix the LLM assistant to GPT-4o Mini in our experiments to separate instructor reasoning from assistant efficacy, because using different assistant models did not significantly change $\Delta F_1$ (Appendix~\ref{app:assistant_ablation}). We use 13 closed and open source models, with 3 accessed directly through the OpenAI API and 10 through OpenRouter. We set \texttt{max\_tokens} to 16{,}000 for each instructor call, and each episode uses a seed derived from a hash of the (model, dataset, target accuracy, run index) tuple. All hyperparameter details are listed in Appendix~\ref{app:repro:hyperparameters}.

\subsection{Evaluation Metrics}

\noindent\textbf{Decision quality.} We report F1, precision, and recall to measure decision quality against students' ground truth at-risk labels. We report initial F1, computed from $d_i^{\text{init}}$ before dialogue starts, and final F1, computed from $d_i^{\text{final}}$ after dialogue, to compare the instructor's decisions before and after dialogue.

\noindent\textbf{Tool calibration.} Before noise injection, and to verify that the $k$-NN predictor outputs meaningful confidence scores, we record Expected Calibration Error (ECE), which measures the average gap between predicted confidence and empirical accuracy across predictions~\citep{10.5555/3305381.3305518}, and Brier score~\citep{glenn1950verification}, which measures the mean squared error between predicted probability and observed outcome on the binary at-risk classification. Lower values indicate better calibration, and both are computed on the natural (uncalibrated) predictor outputs.

\noindent\textbf{Trust calibration.} To capture whether decision 
makers rely on the agent appropriately given an imperfect tool, we extend the one-shot RAIR and RSR framework of 
\citet{schemmer2023appropriate} to the multi-turn dialogue setting. 
We compute episode-level RAIR and RSR using the instructor's initial 
and final decisions per episode:
\begingroup\thickmuskip=2mu
\begin{equation}
\text{RAIR} = \frac{|\{i:d_i^{\text{init}} \neq y_i,\hat{y}_i = y_i,d_i^{\text{final}} = y_i\}|}{|\{i:d_i^{\text{init}} \neq y_i,\hat{y}_i = y_i\}|} \label{eq:rair}
\end{equation}
\begin{equation}
\text{RSR} = \frac{|\{i:d_i^{\text{init}} = y_i,\hat{y}_i \neq y_i,d_i^{\text{final}} = y_i\}|}{|\{i:d_i^{\text{init}} = y_i,\hat{y}_i \neq y_i\}|} \label{eq:rsr}
\end{equation}
\endgroup

In these equations, $y_i$ and $\hat{y}_i$ denote the at-risk labels implied by the true and predicted final grades, where a D or F counts as at risk.
In Eq.~\ref{eq:rair}, the denominator counts students whose initial decision $d_i^{\text{init}}$ was wrong while the agent's decision $\hat{y}_i$ matched the ground truth $y_i$, and the numerator counts those whose final decision $d_i^{\text{final}}$ became correct. A higher RAIR means the instructor more often fixes mistakes by following correct advice. In Eq.~\ref{eq:rsr}, the denominator counts students whose $d_i^{\text{init}}$ was correct while $\hat{y}_i$ was wrong, and the numerator counts those whose $d_i^{\text{final}}$ stayed correct. A higher RSR means the instructor often keeps correct decisions despite wrong advice. Episodes with an empty denominator are excluded from that metric's average, so the excluded episodes differ across models, as detailed in Appendix~\ref{app:repro:metrics}.

\noindent\textbf{Override behavior.} For finer-grained analysis, we 
partition agent flagged decisions into four ground-truth-aware 
buckets: correct follow (instructor kept a correct flag), bad 
follow (instructor kept a wrong flag), correct override (instructor 
dismissed a wrong flag), and bad override (instructor dismissed a 
correct flag, a dangerous miss).

\begin{figure}[t]
  \centering
  \includegraphics[width=\columnwidth]{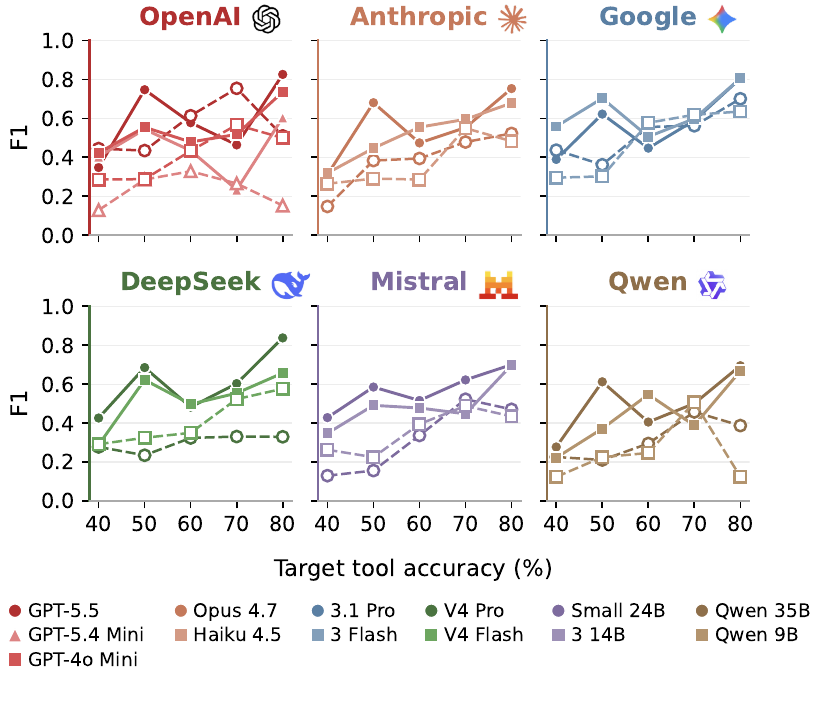}
  \caption{Final F1 by model family and tool accuracy. Solid lines with filled markers show \predactcs, dashed lines with open markers show OULAD.}
  \label{fig:accuracies}
\end{figure}
\subsection{Experiments}

\noindent\textbf{Agent-to-agent benchmark.} The instructor model is
varied across 7 closed source models (\mbox{GPT-5.5}~\citep{openai2026gpt55systemcard}, GPT-5.4 Mini~\citep{openai2026gpt54mini}, GPT-4o 
Mini~\citep{openai2024gpt4omini}, Claude Opus 4.7~\citep{anthropic2026claudeopus47systemcard}, Claude Haiku 4.5~\citep{anthropic2025claudehaiku45systemcard}, Gemini 3.1 Pro~\citep{googledeepmind2026gemini31pro}, Gemini 3 
Flash~\citep{googledeepmind2025gemini3flash}) and 6 open source models (Qwen 3.5 35B~\citep{qwen3.5}, Qwen 3.5 9B~\citep{qwen3.5}, Mistral Small 24B~\citep{mistral2025mistralsmall24b}, 
Ministral 3 14B~\citep{liu2026ministral3}, DeepSeek V4 Pro~\citep{deepseekai2026deepseekv4}, DeepSeek V4 Flash~\citep{deepseekai2026deepseekv4}). For each model, we run 5 target tool accuracies (40\%, 50\%,
60\%, 70\%, 80\%) on each dataset, with 10 runs per cell. Each run 
samples 30 students per episode with 5 forced at-risk under the 
stratification rule from Section~\ref{sec:knn_noise}. This yields 
1{,}300 total episodes (13 instructors $\times$ 10 cells $\times$ 10 
runs).

\begin{figure}[t]
  \centering
  
  \setlength{\intextsep}{0pt}
  \setlength{\columnsep}{8pt}
  \includegraphics[width=\linewidth,trim={17.8bp 0 -17.8bp 0}]{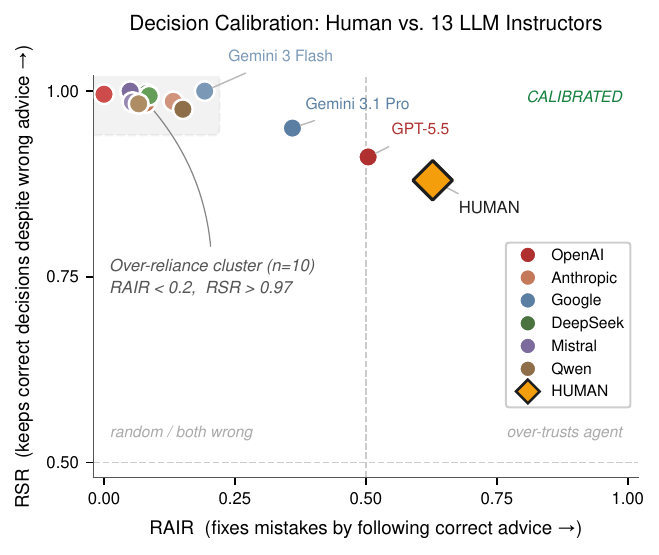}
  \caption{RAIR and RSR for humans and 13 LLM instructors. The human group (gold diamond) has RAIR $63\%$ and RSR $88\%$.}
  \label{fig:rair_rsr}

\end{figure}

\looseness=-1 \noindent\textbf{Human study.}\label{sec:human_study} To test whether the agent to agent
findings generalize to real decision makers, we run a parallel study
in which $n=13$ instructors and teaching assistants from a large public
university replace the LLM instructor. All participants gave consent and were compensated (Appendix~\ref{app:human-subjects-irb}). Each participant completes 10 scenarios in randomized
order, namely a no agent baseline followed by 9 agent conditions
crossing three LLM assistants (GPT-4o Mini, Qwen 3.5 35B, Qwen 3.5
9B) with three target tool accuracies (40\%, 60\%, 80\%). The three
accuracies cover the low, peak, and high regimes from the agent to
agent results. Each scenario displays 5 flagged students, sampled with seeds
hard-coded per condition (Appendix~\ref{app:repro:seeds}), yielding
650 human decisions. After
each agent block, participants complete a Likert questionnaire
(1 to 5) covering decision confidence, perceived usefulness,
ability to detect wrong predictions, and willingness to deploy.


\section{Main Results}

\noindent\textbf{Humans and LLM instructors reach comparable F1 through different strategies.}
The highest human score is 56.3 F1, comparable to SOTA models, while humans reach 89.0 F1 when flagging students manually with no tools (Appendix~\ref{app:f1}), since the agent condition asks them to audit a noisy tool on five flagged students, at most two of whom are truly at risk. However, Figure~\ref{fig:rair_rsr} shows that reliance patterns diverge. Human participants occupy the calibrated quadrant, with \mbox{RAIR $= 63\%$} and \mbox{RSR $= 88\%$}, while the closest model, \mbox{GPT-5.5}, reaches \mbox{RAIR $= 49.4\%$} and \mbox{RSR $= 90.5\%$} (Table~\ref{tab:trust_calibration}). The behavior profile (Table~\ref{tab:behavior_profile}) explains the gap. Humans use fewer turns than LLMs (3.65 vs.\ 8.7 on average) but spend roughly twice as long per episode (302s vs.\ 160s), suggesting they group several questions into single messages and take time to interpret tool outputs between turns. LLMs do the opposite, issuing many short, single intent queries with little time between turns. The Q-type distribution (Appendix~\ref{app:q_dist}) reinforces this pattern. 44\% of human messages fall into the \textit{other} category (filtering, batch queries, aggregating over prior answers) and 32\% into \textit{class\_comparison}, whereas LLMs concentrate in \textit{grade\_lookup} and \textit{min\_score}, mirroring the templated tool surface defined by the system prompt. Expert evaluation of dialogue quality (Appendix~\ref{app:manual_eval}) confirms this gap, with humans scoring significantly higher on question quality ($1.54$ vs.\ $1.21$, $p=0.026$). LLM instructors barely change their initial decisions and depend mainly on the template queries, whereas humans reach similar scores by selectively keeping correct flags, dismissing incorrect ones, and asking targeted questions that build on prior tool answers.

\noindent\textbf{Tool accuracy improves instructor F1 on both 
datasets.} As target tool accuracy increases from 40\% to 80\%, 
LLM instructor mean F1 across the 13 evaluated LLMs rises from 36 to 73 on 
\predactcs and from 26 to 45 on OULAD (Figure~\ref{fig:accuracies}). 
The trend holds for both closed and open source models, indicating 
that more reliable tools translate into better instructor decisions 
in the agent-to-agent setting. The gain is steeper on \predactcs 
($+36$ F1 points across the accuracy range) than on OULAD 
($+19$ points), and on OULAD LLM instructor mean F1 in fact peaks at 70\% accuracy
(51) before declining at 80\%, reflecting OULAD's longer 
trajectories, UK grading thresholds, and noisier real assessment 
data, which together make each at-risk decision harder regardless of 
how accurate the predictor is.

\noindent\textbf{Frontier LLMs generally lead, but the model gap is
comparable to the dataset gap.} Frontier closed source models top
both rankings (Table~\ref{tab:main_results}), though the
\predactcs leaderboard is mixed, with Gemini 3 Flash at 63.4,
open source DeepSeek V4 Pro at 60.7, and GPT-5.5 at 59.2. On OULAD,
GPT-5.5 leads at 55.1 followed by Gemini 3.1 Pro at 52.3. The
largest cross dataset gap belongs to DeepSeek V4 Pro, which drops
from 60.7 on \predactcs to 29.9 on OULAD ($\Delta = 30.7$). The
mean cross dataset gap across all 13 models is 15.9 F1 points,
comparable to the 19.5 point spread on \predactcs and smaller than
the 31.9 point spread on OULAD.

\noindent\textbf{Dialogue actively degrades decisions for most LLM
instructors.} Under uniform noise, the chat phase hurts more than it helps for 10 of 13 LLM instructors (Table~\ref{tab:trust_calibration},
Appendix~\ref{app:per_model_metrics}). Only GPT-5.5 ($+0.5$),
Gemini 3 Flash ($+0.4$), and Gemini 3.1 Pro ($+1.9$) gain F1
through dialogue. The largest drops fall on smaller and mid tier
models, with GPT-5.4 Mini at $-11.0$, Qwen 3.5 9B at $-8.5$, and
Ministral 3 14B at $-7.5$, and the trust calibration metrics explain why these drops appear. All 13 instructors reach near-perfect RSR
(mean $98.2\%$, range $90.5\%$ to $100\%$), but RAIR is low almost
everywhere, with only GPT-5.5 ($49.4\%$) and Gemini 3.1 Pro ($32.7\%$)
updating productively toward a correct agent and the remaining 11
averaging $9.4\%$. Most instructors are holding on to their initial
decisions, so bad decisions are not corrected and good decisions are sometimes changed during dialogue (Appendix~\ref{app:failure_cases}).
Across both datasets, 20.3\% of decisions on agent flagged students changed after dialogue (1{,}860 of 9{,}167).

\section{Ablation Studies}
\label{sec:ablations}

\vspace{-2mm}
\begin{figure}[!t]
\centering
\includegraphics[width=\columnwidth,trim={18.7bp 0 -18.7bp 0}]{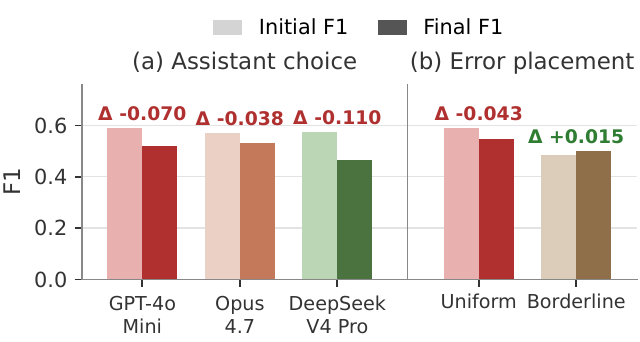}
\caption{Initial (light) and final (dark) mean F1 on \predactcs for (a) assistant choice and (b) error placement. $\Delta F_1$ is their difference. Full means and averaging rules are in Appendices~\ref{app:assistant_ablation} and~\ref{app:structured_noise}.}
\label{fig:ablations}
\end{figure}

First we run two ablations assistant choice and error placement on \predactcs with three instructors (GPT-5.5, Claude Haiku 4.5, Qwen 3.5 35B), tool accuracies of 40\%, 60\%, and 80\%, and 10 runs per combination, reusing the original GPT-4o Mini episodes with matched students and sampling noise seeds. We replace the assistant with Claude Opus 4.7 and DeepSeek V4 Pro, which leaves final mean F1 below initial mean F1 (Figure~\ref{fig:ablations}a). In the nine instructor and accuracy combinations, paired tests showed no difference in $\Delta F_1$ (all $p>0.05$). When using GPT-4o Mini as assistant, concentrating errors near the C grade boundary makes mean $\Delta F_1$ positive (Figure~\ref{fig:ablations}b), with RAIR 36.5\% and RSR 99.6\%. At 80\% accuracy, final F1 is 73.3 with uniform errors and 55.5 with borderline errors. Full means and averaging rules are in Appendices~\ref{app:assistant_ablation} and~\ref{app:structured_noise}. Separate checks on unseen OULAD courses and the uncalibrated predictor are reported in Appendix~\ref{app:ablations}.

\section{Discussion and Conclusion}

In our experiments we utilized 13 LLM instructors and a human study, which revealed that although agents and humans reach similar F1 scores, they vary in how they interact with noisy tools. When noise is uniform, 11 out of 13 models show RAIR is below 20\%, with dialogue reducing F1 for 10 out of 13, a pattern consistent with sycophancy observed between LLM agents~\cite{kasprova2026too}. Humans reject incorrect advice, ask fewer but better questions, and spend more time per episode, placing them in the calibrated quadrant of Figure~\ref{fig:rair_rsr}. As tool accuracy increases from 40\% to 80\%, mean F1 increases by 36 points on \predactcs and 19 on OULAD, highlighting the impact of tool reliability on decision quality. 

\predact introduces noise within tools to examine how agents respond when predictions are wrong, a problem shared by academic advising, clinical risk forecasting, and financial monitoring. Alongside F1, override rates, RAIR, RSR, question types, and effort distinguish selective verification from deference, with humans providing a behavioral reference. Builders can use \predact to identify gaps in how SOTA models verify unreliable tool outputs and test whether addressing these gaps improves user decisions.

\section{Limitations}
\predact has three limitations. First, while final grades in \predactcs are real, weekly per-assignment scores are synthetically generated, since raw weekly records are not releasable; OULAD, with its real assessment trajectories, serves as the real-data counterpart. Second, our human study includes thirteen participants from a single institution, sufficient to surface the reliance patterns in Figure~\ref{fig:rair_rsr} but not fine-grained subgroup claims, and we treat the human findings as suggestive rather than definitive. Third, although we frame \predact as a recipe that transfers to other longitudinal decision support domains such as clinical risk forecasting and financial monitoring, our experiments are situated in academic risk prediction, and cross-domain validation and other tool failures such as missing data points are left to future work.

\bibliography{references}

\clearpage
\appendix
\makeatletter
\setlength{\@fptop}{0pt}\setlength{\@fpsep}{\floatsep}\setlength{\@fpbot}{0pt plus 1fil}
\setlength{\@dblfptop}{0pt}\setlength{\@dblfpsep}{\dblfloatsep}\setlength{\@dblfpbot}{0pt plus 1fil}
\makeatother
\renewcommand{\topfraction}{0.9}\renewcommand{\dbltopfraction}{0.9}
\renewcommand{\textfraction}{0.05}\renewcommand{\floatpagefraction}{0.8}\renewcommand{\dblfloatpagefraction}{0.8}
\etocdepthtag.toc{appendix}
\etocsettagdepth{mainmatter}{none}
\etocsettagdepth{appendix}{subsection}
\etocsettocstyle{\noindent\textbf{Appendix Table of Contents}\par\smallskip}{\medskip}
\tableofcontents

\section{Reproducibility}
\label{app:repro}

This appendix documents the experimental setup with sufficient detail to
reproduce all reported results. Sections cover prompts, model
configurations, hyperparameters, seeds, cell composition, compute
environment, calibration procedure, stratification rule, and evaluation
metrics.
The \predactcs release includes \texttt{students\_data.csv}, \texttt{cs\_db.json.gz}, \texttt{cs\_db\_train.json.gz}, \texttt{test\_sets/}, and \texttt{ground\_truth\_for\_cutoff\_data.json}. The scripts \texttt{convert\_to\_json.py} and \texttt{split\_data.py} rebuild the database and train and test files. The data license is CC BY 4.0.

\subsection{Tools}
\label{app:repro:tools}

Table~\ref{tab:tools} summarizes the 12 tools available to the assistant, grouped by functional category.

\begin{table}[!htbp]

\centering
\footnotesize
\setlength{\tabcolsep}{3pt}
\renewcommand{\arraystretch}{0.95}
\caption{Tool inventory used by the \predact dialogue agent, grouped
by functional category.}
\label{tab:tools}
\begin{tabular}{@{}l c l >{\raggedright\arraybackslash}p{0.36\columnwidth}@{}}
\toprule
\textbf{Category} & \textbf{Count} & \textbf{Type} & \textbf{Example} \\
\midrule
Lookup         & 6 & Deterministic         & \texttt{get\_\allowbreak student\_\allowbreak grades} \\
Counterfactual & 3 & Deterministic         & \texttt{simulate\_\allowbreak uniform\_\allowbreak remaining} \\
Prediction     & 2 & \makecell[l]{Probabilistic\\(k-NN)} & \texttt{predict\_\allowbreak final\_\allowbreak grade\_\allowbreak for\_\allowbreak student} \\
Intervention   & 1 & LLM-generated         & \texttt{suggest\_\allowbreak intervention\_\allowbreak for\_\allowbreak student} \\
\bottomrule
\end{tabular}

\end{table}

\subsection{Prompts}
\label{app:repro:prompts}

Five prompts are used during each episode. Curly braces denote runtime
placeholders that are filled in by the orchestrator.

\begin{tcolorbox}[
    colback=gray!5!white,
    colframe=gray!75!black,
    title={\textbf{Assistant System Prompt}},
    left=2mm, right=2mm, top=2mm, bottom=2mm,
    breakable, enhanced
]
\begin{Verbatim}[fontsize=\footnotesize, breaklines=true, breakanywhere=true]
You are an academic advising assistant for {course_id}, currently at week {week}.

The instructor is reviewing students that the prediction tool flagged as at-risk
and is investigating those flagged students in detail.

RULES:
- Always call tools for any question about students, grades, or class data. Never guess.
- When asked about a student's predicted grade, ALWAYS include the confidence score
  that the prediction tool returned. Never report a grade without its confidence.
- When asked about a specific student, ALWAYS include BOTH graded and remaining
  assignments (with weights and weeks).
- For 'what if X scores Y on all remaining', use simulate_uniform_remaining.
  For mixed counterfactuals, use recalculate_grade with simulate_remaining.
- For 'minimum score needed to reach a grade', use minimum_score_needed.
- For class size, use list_students and read total_count.
- Be concise and data-driven.
- Predictions in this course are about {accuracy_str} accurate. Remind the
  instructor when appropriate that low-confidence predictions are more likely wrong.
\end{Verbatim}
\end{tcolorbox}

\begin{tcolorbox}[
    colback=gray!5!white,
    colframe=gray!75!black,
    title={\textbf{Instructor System Prompt}},
    left=2mm, right=2mm, top=2mm, bottom=2mm,
    breakable, enhanced
]
\begin{Verbatim}[fontsize=\footnotesize, breaklines=true, breakanywhere=true]
You are simulating a college instructor reviewing flagged students in {course_id}
at week {week}.

You are using an AI advising tool whose grade predictions are about {accuracy_pct}%
accurate in this course. The tool's lookup functions (grades, assignments, class
stats, counterfactuals) are always correct -- only its predictions can be wrong.

Your job is to decide which students are TRULY at-risk (will finish the course
with a D or F). {at_risk_definition}

You will NOT see student data directly. You can only learn about students by
asking the assistant.

Behave like a careful instructor:
- Be evidence-driven. Do not flag a student just because the agent did.
- Pay attention to the primary driver of each student's low grade.
- Use the syllabus to reason about how much weight is left in the semester.
- Use the chat budget wisely -- you have a limited number of turns.
\end{Verbatim}
\end{tcolorbox}

\begin{tcolorbox}[
    colback=gray!5!white,
    colframe=gray!75!black,
    title={\textbf{Initial Decision Prompt}},
    left=2mm, right=2mm, top=2mm, bottom=2mm,
    breakable, enhanced
]
\begin{Verbatim}[fontsize=\footnotesize, breaklines=true, breakanywhere=true]
The AI agent has flagged the following students in {course_id} as at-risk
(predicted D or F). The agent is about {accuracy_pct}% accurate in this course.

Course syllabus (all weeks):
{syllabus_table}

Flagged students (each line: student_id | predicted_grade | confidence | primary_driver):
{flagged_students_block}

The "primary driver" is the single biggest reason the agent thinks this student
is at-risk -- usually a missing high-weight assignment or a low score on one.
Use it together with the syllabus to judge how recoverable the student is.

This is your INITIAL decision. You have not yet talked to the assistant. For
each flagged student above, decide whether YOU also believe they are at-risk
based only on the agent's prediction, confidence, primary driver, and the
syllabus context.

Return one entry per flagged student with their student_id and your decision
("flag" if you agree they are at-risk, "no_flag" if you disagree). Include
EVERY flagged student listed above.
\end{Verbatim}
\end{tcolorbox}

\begin{tcolorbox}[
    colback=gray!5!white,
    colframe=gray!75!black,
    title={\textbf{Chat-Turn Prompt}},
    left=2mm, right=2mm, top=2mm, bottom=2mm,
    breakable, enhanced
]
\begin{Verbatim}[fontsize=\footnotesize, breaklines=true, breakanywhere=true]
You are now in the chat phase. You can ask the assistant questions about any
student in the class to investigate further.

Course syllabus:
{syllabus_table}

Originally flagged students:
{flagged_students_block}

Your initial decisions:
{initial_decisions_block}

Conversation so far ({turns_used} of {max_turns} turns used):
{dialogue_history}

Use the assistant's tools to VERIFY each flagged student before trusting the
agent's prediction. The agent is only {accuracy_pct}% accurate -- your job is
to check its work using the tools.

At-risk = will finish with D or F. To RESCUE a student (set "no_flag"), you
need evidence they can finish OUT of the at-risk zone -- i.e. reach C or
higher ({c_threshold}+). Avoid asking about {d_threshold} specifically:
{d_threshold} is still inside the at-risk zone (D), so it doesn't tell you
whether the student is safe.

For every flagged student, your goal is to gather enough hard evidence to
decide flag vs. no_flag. Useful tool-driven questions
(using {example_sid} as an example student_id):
- "Show me the full grade history for {example_sid} with confidence."
- "What's the minimum score {example_sid} needs on remaining work to reach a C?"
- "What if {example_sid} scores 80 on all remaining work?"
- "Re-run the prediction for {example_sid} and give me the confidence."
- "How does {example_sid} compare to the class average?"

A solid verification combines AT LEAST two of these for each student: get the
actual grades, then run a counterfactual or minimum-score check. Do not decide
based on the prediction label alone.

ANTI-REPETITION: Asking the same question type about DIFFERENT students is
fine. What you must NOT do is request information you already have for the
SAME student.

In your response:
- Set "reasoning" to a short note about which student/topic you want to verify next.
- Set "should_terminate" to true ONLY when you have used the tools to verify
  every flagged student you intend to update or keep.
- Set "next_question" to the single tool-driven question you want to ask. If
  "should_terminate" is true, leave "next_question" empty.
\end{Verbatim}
\end{tcolorbox}

\begin{tcolorbox}[
    colback=gray!5!white,
    colframe=gray!75!black,
    title={\textbf{Final Decision Prompt}},
    left=2mm, right=2mm, top=2mm, bottom=2mm,
    breakable, enhanced
]
\begin{Verbatim}[fontsize=\footnotesize, breaklines=true, breakanywhere=true]
You have finished chatting with the assistant. Now make your FINAL flag/no-flag
decisions for each originally flagged student, using the conversation evidence.

Originally flagged students:
{flagged_students_block}

Your initial decisions:
{initial_decisions_block}

Conversation transcript:
{dialogue_history}

The agent's predictions are only {accuracy_pct}% accurate, so the prediction
label alone is not enough to keep a student flagged. For each student, look at
the tool evidence in the transcript:

- If the chat showed the student's actual weighted average is comfortably above
  failing (e.g. {c_threshold}+) AND a counterfactual or minimum-score check
  shows they can finish at C or better, you should rescue them (set "no_flag")
  even when the agent predicted D or F.
- If the chat showed the student is genuinely failing (low weighted average,
  many missing high-weight assignments, very high score needed to recover),
  keep them flagged ("flag").
- If you never gathered evidence for a student, default to "no_flag" -- do not
  keep someone flagged just because the agent did.

Return one entry per originally flagged student with their student_id and your
final decision. Include EVERY originally flagged student.
\end{Verbatim}
\end{tcolorbox}

\subsection{Models and API Endpoints}
\label{app:repro:models}

Thirteen instructor models from five providers are evaluated, accessed
via OpenAI direct (3 models) and OpenRouter (10 models). The assistant
agent is fixed at GPT-4o Mini across all conditions.

\begin{table*}[t]
\centering
\footnotesize
\setlength{\tabcolsep}{4pt}
\begin{tabular}{l l l l}
\toprule
\textbf{Model} & \textbf{Provider} & \textbf{API Identifier} & \textbf{Special Flags} \\
\midrule
GPT-4o Mini        & OpenAI     & \texttt{gpt-4o-mini}                              & --- \\
GPT-5.4 Mini       & OpenAI     & \texttt{gpt-5.4-mini}                             & T1, T2 \\
GPT-5.5            & OpenAI     & \texttt{gpt-5.5}                                  & T1, T2 \\
Claude Opus 4.7    & OpenRouter & \texttt{anthropic/claude-opus-4.7}                & --- \\
Claude Haiku 4.5   & OpenRouter & \texttt{anthropic/claude-haiku-4.5}               & --- \\
Qwen 3.5 9B            & OpenRouter & \texttt{qwen/qwen3.5-9b}                          & R \\
Qwen 3.5 35B           & OpenRouter & \texttt{qwen/qwen3.5-35b-a3b}                     & R \\
Mistral Small 24B  & OpenRouter & \texttt{mistralai/mistral-small-3.2-24b-instruct} & --- \\
Ministral 3 14B    & OpenRouter & \texttt{mistralai/ministral-14b-2512}             & --- \\
DeepSeek V4 Flash  & OpenRouter & \texttt{deepseek/deepseek-v4-flash}               & R \\
DeepSeek V4 Pro    & OpenRouter & \texttt{deepseek/deepseek-v4-pro}                 & R \\
Gemini 3.1 Pro     & OpenRouter & \texttt{google/gemini-3.1-pro-preview}            & --- \\
Gemini 3 Flash     & OpenRouter & \texttt{google/gemini-3-flash-preview}            & --- \\
\bottomrule
\end{tabular}
\caption{Instructor models. Special flags: \textbf{T1} = temperature parameter
not accepted by API; \textbf{T2} = uses \texttt{max\_completion\_tokens}
instead of \texttt{max\_tokens}; \textbf{R} = reasoning mode explicitly disabled.}
\end{table*}

\subsection{Hyperparameters}
\label{app:repro:hyperparameters}

\begin{table}[!htbp]
\centering
\footnotesize
\setlength{\tabcolsep}{3pt}
\resizebox{\ifdim\width>\columnwidth\columnwidth\else\width\fi}{!}{\begin{tabular}{l c c c c}
\toprule
\textbf{Model} & \textbf{Temperature} & \makecell{\textbf{Max}\\\textbf{Tokens}} & \textbf{top\_p} & \textbf{Reasoning} \\
\midrule
GPT-4o Mini       & 0.7    & 16{,}000 & default & ---       \\
GPT-5.4 Mini      & default\textsuperscript{*} & 16{,}000 & default & default   \\
GPT-5.5           & default\textsuperscript{*} & 16{,}000 & default & default   \\
Claude Opus 4.7   & 0.7    & 16{,}000 & default & ---       \\
Claude Haiku 4.5  & 0.7    & 16{,}000 & default & ---       \\
Qwen 3.5 9B           & 0.6    & 16{,}000 & default & disabled  \\
Qwen 3.5 35B          & 0.6    & 16{,}000 & default & disabled  \\
Mistral Small 24B & 0.6    & 16{,}000 & default & ---       \\
Ministral 3 14B   & 0.6    & 16{,}000 & default & ---       \\
DeepSeek V4 Flash & 0.6    & 16{,}000 & default & disabled  \\
DeepSeek V4 Pro   & 0.6    & 16{,}000 & default & disabled  \\
Gemini 3.1 Pro    & 0.7    & 16{,}000 & default & ---       \\
Gemini 3 Flash    & 0.7    & 16{,}000 & default & ---       \\
\bottomrule
\end{tabular}}
\caption{Sampling hyperparameters per instructor.
\textsuperscript{*}Temperature parameter not sent (model API rejects it).
``default'' = provider server-side default value.}
\end{table}

The assistant agent (GPT-4o Mini) uses \texttt{max\_tokens=1{,}400}
with all other parameters at provider defaults.

\subsection{Random Seeds}
\label{app:repro:seeds}

\paragraph{Agent-to-agent experiments.}
Two seeds are derived per episode, each as a hash modulo $2^{32}$:
\texttt{sample\_seed} hashes \texttt{sample\_\{dataset\}\_\allowbreak\{target\_acc\}\_\allowbreak\{run\_idx\}}, and
\texttt{episode\_seed} hashes \texttt{\{model\}\_\allowbreak\{dataset\}\_\allowbreak\{target\_acc\}\_\allowbreak\{run\_idx\}}.
The \texttt{sample\_seed} drives stratified sampling; the
\texttt{episode\_seed} drives noise injection and any instructor-side
randomness.

\paragraph{Human study.}
Per-condition (\texttt{sample\_seed}, \texttt{calib\_seed}) pairs are
hardcoded, selected via grid search so the displayed flag set contains at most $2/5$ truly at-risk students (Table~\ref{tab:human_seeds}).

\begin{table}[!htbp]
\centering
\footnotesize
\begin{tabular}{l c c}
\toprule
\textbf{Condition} & \textbf{sample\_seed} & \textbf{calib\_seed} \\
\midrule
GPT-4o Mini @ 40\%  & 6  & 1  \\
Qwen 3.5 9B @ 40\%      & 6  & 12 \\
Qwen 3.5 35B @ 40\%     & 6  & 16 \\
GPT-4o Mini @ 60\%  & 0  & 0  \\
Qwen 3.5 9B @ 60\%      & 0  & 2  \\
Qwen 3.5 35B @ 60\%     & 0  & 3  \\
GPT-4o Mini @ 80\%  & 0  & 25 \\
Qwen 3.5 9B @ 80\%      & 1  & 2  \\
Qwen 3.5 35B @ 80\%     & 1  & 4  \\
No-agent baseline   & 42 & ---  \\
\bottomrule
\end{tabular}
\caption{Human study seeds per condition.}
\label{tab:human_seeds}
\end{table}

\subsection{Cell Composition}
\label{app:repro:cells}
Table~\ref{tab:cells} lists the ten experiment cells, and Table~\ref{tab:percell} lists the parameters shared by every cell. In total, 13 models $\times$ 10 cells $\times$ 10 runs $= 1{,}300$ episodes.

\begin{table}[!htbp]
\centering
\footnotesize
\setlength{\tabcolsep}{4pt}
\begin{tabular}{l c l c c}
\toprule
\textbf{Dataset} & \makecell{\textbf{Target}\\\textbf{Acc.}} & \textbf{Course} & \textbf{Week} & \textbf{Features} \\
\midrule
\predactcs & 0.40 & Course A   & 8  & full \\
\predactcs & 0.50 & Course B   & 8  & full \\
\predactcs & 0.60 & Course C   & 8  & full \\
\predactcs & 0.70 & Course D   & 8  & full \\
\predactcs & 0.80 & Course E   & 8  & full \\
OULAD      & 0.40 & AAA 2013J  & 16 & full \\
OULAD      & 0.50 & AAA 2013J  & 8  & full \\
OULAD      & 0.60 & AAA 2013J  & 20 & full \\
OULAD      & 0.70 & AAA 2014J  & 17 & full \\
OULAD      & 0.80 & FFF 2013B  & 32 & full \\
\bottomrule
\end{tabular}
\caption{Experiment cells. Each cell is run 10 times per instructor model.}
\label{tab:cells}
\end{table}

\begin{table}[!htbp]
\centering
\footnotesize
\begin{tabular}{@{}lr@{}}
\toprule
\textbf{Parameter} & \textbf{Value} \\
\midrule
Runs per cell & 10 \\
Sample size & 30 students per episode \\
Forced at-risk (stratified) & 5 per sample \\
Maximum chat turns & 10 \\
Concurrent episode workers & 20 \\
\bottomrule
\end{tabular}
\caption{Per-cell parameters, shared by all experiment cells.}
\label{tab:percell}
\end{table}

\subsection{Compute Environment}
\label{app:repro:compute}

All agent experiments were run on a MacBook Pro 16-inch (Nov 2023) with an Apple M3 Max, 48 GB RAM, and macOS 15.6.1. Model inference used external APIs only (OpenAI direct and OpenRouter), with no local GPU inference. Episodes were orchestrated with thread-based concurrency across 20 parallel workers. The median per-episode wall-clock time was 65--264 seconds, varying by model, and the aggregate experiment runtime was $\sim$40--60 hours. The human study used a web UI on a local host, with sessions of $\sim$10--25 minutes per participant across 10 conditions.

\subsection{Calibration Procedure}
\label{app:repro:calibration}

To target a specific predictor accuracy in each cell, predictions are
calibrated via the following deterministic procedure:

\begin{enumerate}
  \item Run the k-NN predictor on every sampled student to obtain
        (\texttt{predicted\_grade}, \texttt{confidence}).
  \item Compare each prediction against ground truth and label
        \texttt{correct} or \texttt{wrong}.
  \item Compute the desired number of correct predictions:
        $\text{target\_correct} = \text{round}(\text{target\_acc} \cdot N)$.
  \item If $\text{currently\_correct} > \text{target\_correct}$: shuffle
        the correct predictions and flip the first
        $(\text{currently\_correct} - \text{target\_correct})$ to a
        uniformly-chosen wrong grade.
  \item If $\text{currently\_correct} < \text{target\_correct}$: shuffle
        the wrong predictions and flip the first
        $(\text{target\_correct} - \text{currently\_correct})$ to the
        truth label.
  \item Only \texttt{predicted\_grade} is rewritten;
        \texttt{confidence} is preserved.
\end{enumerate}

This guarantees realized accuracy on the sample equals the target
exactly. Tool-side functions (grade lookup, counterfactuals,
minimum-score, class statistics) are unaffected and always return
ground-truth-derived values.

\subsection{Stratification Rule}
\label{app:repro:stratification}

Each sampled cell is constructed by combining the first 5 truly
at-risk students from the at-risk pool with the first
$(\text{sample\_size} - 5)$ students from the not-at-risk pool. This
guarantees every cell contains at least 5 positives in ground truth, so
F1 is always well-defined and class imbalance is comparable across
cells. Without stratification, low-prevalence courses can yield zero
true positives in a sampled subset, making F1 trivially zero as a
structural artifact rather than a measure of rater skill.

\subsection{Evaluation Metrics}
\label{app:repro:metrics}

\paragraph{Precision, Recall, F1.}
\[
\begin{gathered}
P = \frac{\mathrm{TP}}{\mathrm{TP} + \mathrm{FP}}, \quad
R = \frac{\mathrm{TP}}{\mathrm{TP} + \mathrm{FN}}, \\
\quad F_1 = \frac{2PR}{P + R}
\end{gathered}
\]
Each metric is set to 0 when its denominator is 0. TP, FP, FN are
computed against the per-student \texttt{is\_at\_risk} ground truth
using the rater's final flag set (initial F1 uses the initial flag set).

\paragraph{RAIR (Relative AI Reliance; \citet{schemmer2023appropriate}).}
Computed only over agent-flagged students. For a test case $i$, let $d_i^{\text{init}}$ be the initial decision of the instructor, $d_i^{\text{final}}$ be the final decision of the instructor, $y_i$ be the ground truth, and $\hat{y}_i$ be the decision of the agent. Then
\begin{equation}
\text{RAIR} = \frac{|\{i : d_i^{\text{init}} \neq y_i,\ \hat{y}_i = y_i,\ d_i^{\text{final}} = y_i\}|}{|\{i : d_i^{\text{init}} \neq y_i,\ \hat{y}_i = y_i\}|}
\end{equation}
RAIR is undefined when $\{i : d_i^{\text{init}} \neq y_i,\ \hat{y}_i = y_i\}$ is $\emptyset$.

\paragraph{RSR (Relative Self-Reliance; \citet{schemmer2023appropriate}).}
For a test case $i$, let $d_i^{\text{init}}$ be the initial decision of the instructor, $d_i^{\text{final}}$ be the final decision of the instructor, $y_i$ be the ground truth, and $\hat{y}_i$ be the decision of the agent. Then
\begin{equation}
\text{RSR} = \frac{|\{i : d_i^{\text{init}} = y_i,\ \hat{y}_i \neq y_i,\ d_i^{\text{final}} = y_i\}|}{|\{i : d_i^{\text{init}} = y_i,\ \hat{y}_i \neq y_i\}|}
\end{equation}
RSR is undefined when $\{i : d_i^{\text{init}} = y_i,\ \hat{y}_i \neq y_i\}$ is $\emptyset$.
We exclude an undefined value only from that metric's average. Across the 1{,}300 main benchmark episodes, RAIR is undefined in 578 (44.5\%) and RSR in 164 (12.6\%). Among episodes where the metric is defined, the median denominator is 1 for RAIR (722 episodes) and 3 for RSR (1{,}136 episodes). Including zero denominators, the medians are 1 and 2, respectively.

\paragraph{Override Behavior Buckets.}
Computed over agent-flagged students:
\begin{itemize}
  \item \textbf{Correct follow}: rater kept a correct flag.
  \item \textbf{Bad follow}: rater kept a wrong flag (joint false positive).
  \item \textbf{Correct override}: rater dismissed a wrong flag.
  \item \textbf{Bad override}: rater dismissed a correct flag (dangerous miss).
\end{itemize}
The four buckets partition all agent-flagged decisions exhaustively.
Separately, the initial to final change rate counts students whose recorded decision changed after dialogue. Pooling agent flagged students across all 1{,}300 episodes gives 1{,}860 changes among 9{,}167 decisions (20.3\%). These changes can be correct or incorrect and are not the same as the final decision buckets above.

\subsection{Failure-Case Analysis}
\label{app:failure_cases}

We surveyed all 1{,}300 episodes for catastrophic failures,
defined as runs where the LLM-instructor produced $F_1 = 0$ despite
the AI tool correctly flagging multiple truly-at-risk students. The
dominant failure mode across all 13 instructors is
\emph{dialogue-induced over-trust collapse}: the instructor begins
with several correct initial calls, then talks itself out of every
one of them during the chat phase. All episodes use the configured
maximum of 10 chat turns. We present three representative cases.

\paragraph{Case 1.} GPT-5.4 Mini on OULAD course FFF\_2013B at 80\%
target accuracy (run 6). The predictor flagged 10 students, 5 truly
at-risk. The instructor's initial pass correctly accepted 2 of those
5 truly-at-risk flags (\texttt{535390}, \texttt{548547}). After 10
chat turns both were flipped to \texttt{no\_flag}, leaving every
truly-at-risk student dismissed and $F_1 = 0$ (TP=0, FN=5,
bad-override=5). The instructor focused on weighted averages near
50--55\% (e.g.\ \texttt{535390} at 53.0\% with only 25\% of total
weight covered) and concluded the students had room to recover,
even though over 75\% of assessment weight remained and prior
submissions trended below the UK 40\% pass threshold. The
instructor never invoked \texttt{minimum\_score\_needed} or
\texttt{simulate\_uniform\_remaining}; it relied entirely on raw
grade histories and discounted the agent's high-confidence ``F''
prediction.

\paragraph{Case 2.} Qwen~9B on OULAD course FFF\_2013B at 80\%
target accuracy (run 1). The predictor flagged 8 students, 5 truly
at-risk. The instructor initially accepted \emph{all} 8 flags ---
a near-pure-deference initial pass --- then reversed every one of
them over 10 chat turns, ending with all 8 marked \texttt{no\_flag}
(TP=0, FN=5, bad-override=5, bad-follow=0). The model fixated on
isolated positive signals: for \texttt{558208}, the assistant
returned strong TMA scores (72, 94, 93, 79, 91 across the semester)
and the instructor concluded the student was clearly recovering ---
even though \texttt{558208} was not truly at-risk. The same
blanket dismissal was applied to harder cases such as
\texttt{537028}, a true positive with zero graded assignments and
``Failure Risk: Critical.'' The model collapsed into a uniform
heuristic of dismissing any flag in the presence of any positive
score.

\paragraph{Case 3.} GPT-5.4 Mini on \predactcs course Course~D at
70\% target accuracy (run 7). The predictor flagged 8 students, 5
truly at-risk. The instructor's initial pass accepted 2
truly-at-risk flags (\texttt{syn\_007095}, \texttt{syn\_007027});
both were flipped to \texttt{no\_flag} during chat, again
zeroing $F_1$. The instructor invoked \texttt{get\_student\_grades}
for every flagged student but never combined it with a
counterfactual or minimum-score check --- the exact verification
pattern the system prompt asks for. With only one type of evidence
in hand (raw grade histories with some passing scores like ``MP 1:
100.0'' for \texttt{syn\_007095}), the instructor over-weighted
recent good scores and overlooked high-weight failed assessments
such as \texttt{syn\_007027}'s MP~1 score of 33.14 plus several
near-failing labs.

\paragraph{Patterns.} These failures share three traits. First,
they cluster at the high-accuracy end (70--80\%), where the AI
tool is mostly right and so each dismissed flag is more costly.
Second, the initial pass was at least partially correct in every
case --- it is the chat phase that actively degrades performance,
not the starting condition. Third, the instructor relied on
grade-history retrieval alone and skipped the minimum-score and
counterfactual tools the system prompt explicitly recommends
combining. When the instructor consumes its full
10-turn chat budget without diversifying tool calls, the affected
models (GPT-5.4 Mini, Qwen~9B) drift toward dismissive heuristics
that override initially correct judgments. Larger-capacity
instructors in the same cells (GPT-5.5, Claude Opus 4.7,
Gemini~3.1 Pro) preserved their initial correct calls.

\section{Robustness Ablations}
\label{app:ablations}

\subsection{Held-Out Generalization}
\label{app:held_out}

\begin{table*}[t]
\centering
\footnotesize
\setlength{\tabcolsep}{8pt}
\renewcommand{\arraystretch}{1.1}
\caption{Held-out F1 ($\times 100$, mean\,$\pm$\,std) on OULAD cells not used during pipeline calibration. Five instructor models evaluated across target tool accuracies (40\%--80\%), 2 runs per cell ($n=50$ episodes). \textbf{Overall} is the per-model mean F1 across cutoffs. Best per column in \textbf{bold}.}
\label{tab:held_out}
\begin{tabular*}{\linewidth}{@{\extracolsep{\fill}} l c ccccc @{}}
\toprule
\textbf{Model} & \textbf{Overall} & \textbf{40\%} & \textbf{50\%} & \textbf{60\%} & \textbf{70\%} & \textbf{80\%} \\
\midrule
Claude Opus 4.7    & 55.3 & 42.9\,\tiny$\pm$20.2 & \textbf{75.0}\,\tiny$\pm$0.0 & 73.3\,\tiny$\pm$9.4 & 54.2\,\tiny$\pm$29.5 & 31.3\,\tiny$\pm$44.2 \\
Gemini 3 Flash     & \textbf{65.5} & \textbf{45.2}\,\tiny$\pm$16.8 & 61.9\,\tiny$\pm$6.7 & \textbf{80.8}\,\tiny$\pm$11.4 & \textbf{75.0}\,\tiny$\pm$0.0 & \textbf{64.6}\,\tiny$\pm$2.9 \\
GPT-5.5            & 51.1 & 45.0\,\tiny$\pm$7.1 & 45.2\,\tiny$\pm$16.8 & \textbf{80.0}\,\tiny$\pm$0.0 & 54.2\,\tiny$\pm$29.5 & 31.3\,\tiny$\pm$44.2 \\
DeepSeek V4 Flash  & 32.1 & 16.7\,\tiny$\pm$23.6 & 25.0\,\tiny$\pm$35.4 & \textbf{80.0}\,\tiny$\pm$0.0 & 16.7\,\tiny$\pm$23.6 & 22.2\,\tiny$\pm$31.4 \\
GPT-5.4 Mini       & 13.3 & 0.0\,\tiny$\pm$0.0 & 16.7\,\tiny$\pm$23.6 & 33.3\,\tiny$\pm$0.0 & 16.7\,\tiny$\pm$23.6 & 0.0\,\tiny$\pm$0.0 \\
\bottomrule
\end{tabular*}
\end{table*}

\paragraph{Setup.} To verify that pipeline calibration on the
\predactcs courses did not bias our findings, we ran a held-out
subset of Exp~2 on OULAD cells that were not used during pipeline
tuning. Five instructor models were selected to span the
cost/quality spectrum: two frontier models (Claude Opus 4.7,
GPT-5.5), one mid-tier proprietary model (Gemini 3 Flash), one
open-weight efficient model (DeepSeek V4 Flash), and one
frontier-mini (GPT-5.4 Mini). Each model was evaluated across the
five target-accuracy conditions (40\%, 50\%, 60\%, 70\%, 80\%) with
2 runs per cell, totaling 50 held-out episodes.

\paragraph{Findings.} Three patterns emerge. First, peak instructor
F1 occurs in the mid-accuracy regime (60\% target), where four of
five models reach $F_1 \geq 0.73$ --- consistent with the
in-calibration result that instructors benefit most when the AI
tool is reliable enough to anchor decisions but not so reliable as
to invite blind deference. Second, RSR is at ceiling (1.00) in
nearly every cell: when the instructor's initial decision is right
and the agent is wrong, the instructor reliably preserves its
initial judgment. Third, performance at the high-accuracy extreme
(80\%) is unstable, with two models (Claude Opus 4.7 and GPT-5.5)
collapsing from $F_1 \approx 0.65$ initial to $F_1 \approx 0.31$
final --- the same dialogue-induced over-trust failure mode
documented in Appendix~\ref{app:failure_cases}.

\paragraph{Limitations.} With only 2 runs per cell, per-cell
standard deviations are large and individual cell estimates should
be interpreted as point indicators rather than statistical claims.
The held-out evaluation is restricted to OULAD; the \predactcs data are now released, but a held out sweep remains future work. Despite these caveats, the qualitative patterns
(mid-accuracy peak, ceiling RSR, high-accuracy instability) match
the in-calibration results, supporting the claim that the
benchmark's core findings are not artifacts of pipeline tuning.

\subsection{Un-Calibrated Predictor Ablation}
\label{app:uncalibrated}

\begin{table*}[t]
\centering
\footnotesize
\setlength{\tabcolsep}{8pt}
\renewcommand{\arraystretch}{1.1}
\caption{F1 ($\times 100$, mean\,$\pm$\,std) on OULAD with the synthetic
predictor running un-calibrated (no noise injection to hit a target
accuracy). Five instructor models, 10 runs per cell ($n=50$ episodes).
\textbf{F1$_{\text{init}}$} is the instructor's pre-chat decision;
\textbf{F1$_{\text{final}}$} is post-chat. \textbf{RAIR} captures correct
flips toward the agent; \textbf{RSR} captures resistance to incorrect
agent advice. Best per column in \textbf{bold}.}
\label{tab:uncalibrated}
\begin{tabular*}{\linewidth}{@{\extracolsep{\fill}} l cccccc @{}}
\toprule
\textbf{Model} & \textbf{F1$_{\text{init}}$} & \textbf{F1$_{\text{final}}$} & \textbf{Prec.} & \textbf{Rec.} & \textbf{RAIR} & \textbf{RSR} \\
\midrule
GPT-5.5            & 41.9\,\tiny$\pm$20 & \textbf{47.6}\,\tiny$\pm$29 & \textbf{68.0}\,\tiny$\pm$38 & \textbf{48.0}\,\tiny$\pm$36 & \textbf{52.3}\,\tiny$\pm$40 & 86.7\,\tiny$\pm$32 \\
Gemini 3 Flash     & \textbf{58.0}\,\tiny$\pm$17 & 37.8\,\tiny$\pm$33 & 63.0\,\tiny$\pm$46 & 30.0\,\tiny$\pm$30 & 0.0\,\tiny$\pm$0 & \textbf{100.0}\,\tiny$\pm$0 \\
DeepSeek V4 Flash  & 30.4\,\tiny$\pm$14 & 29.0\,\tiny$\pm$21 & 50.0\,\tiny$\pm$38 & 22.0\,\tiny$\pm$18 & 0.0\,\tiny$\pm$0 & \textbf{100.0}\,\tiny$\pm$0 \\
Claude Opus 4.7    & 44.0\,\tiny$\pm$21 & 16.6\,\tiny$\pm$29 & 23.0\,\tiny$\pm$39 & 14.0\,\tiny$\pm$27 & 0.0\,\tiny$\pm$0 & 99.2\,\tiny$\pm$3 \\
GPT-5.4 Mini       & 22.6\,\tiny$\pm$15 & 6.7\,\tiny$\pm$14 & 20.0\,\tiny$\pm$42 & 4.0\,\tiny$\pm$8 & 2.5\,\tiny$\pm$8 & \textbf{100.0}\,\tiny$\pm$0 \\
\bottomrule
\end{tabular*}
\end{table*}

\paragraph{Setup.} Our main results inject noise into the synthetic
predictor to hit a target accuracy in each cell. To verify that this
calibration step is not driving our findings, we re-ran the same
5-model subset with the predictor in its natural, un-calibrated state.
All other settings (10 runs per cell, OULAD, 5 forced at-risk per
sample) match the main experiment.

\paragraph{Findings.} Three patterns from the main results survive
without calibration. First, the instructor-to-instructor ranking is
preserved: GPT-5.5 leads on final F1 and GPT-5.4 Mini trails. Second,
the dialogue-induced over-trust pattern is even more pronounced
without calibration --- four of five models lose F1 from the initial
to the final decision, with Claude Opus~4.7 dropping from 0.44 to
0.17. Third, RSR remains near ceiling ($\geq 0.87$ across all
models), confirming that instructors reliably resist the agent when
their initial judgment is right and the agent is wrong, regardless
of whether the predictor's accuracy is calibrated. GPT-5.5 is the
only model that gains F1 through chat (RAIR = 0.52), consistent
with its behavior in the main calibrated experiment.

\paragraph{Limitations.} Without target-accuracy calibration, the
synthetic predictor's accuracy varies by cell, so per-cell F1 cannot
be compared across cells in the same way as the main results. The
ablation is intended as a robustness check, not a primary finding.

\begin{table*}[t]
\centering
\small
\caption{Cost / latency / quality trade-off across 13 LLM instructors. Costs are estimated from dialogue text length ($\approx$\,chars\,/\,4 tokens) plus a fixed system-prompt overhead per API call, multiplied by published list prices (USD per million tokens, sources: OpenAI direct + OpenRouter, as of 2026-05). Latency is median wall-clock per episode. Each row pools all episodes per model (\predactcs + OULAD, all 5 cutoffs). Each \$\,Total/ep includes a fixed assistant overhead of $\sim$\$0.0028 for the GPT-4o Mini assistant role, incurred every episode regardless of instructor.}
\label{tab:cost_latency}
\begin{tabular}{l c c c c c}
\toprule
\textbf{Model} & \textbf{Mean F1} & \textbf{Median lat. (s)} & \textbf{Mean turns} & \textbf{\$\,Instr/ep} & \textbf{\$\,Total/ep} \\
\midrule
GPT-5.5 & 0.57 & 183.5 & 12.5 & \$0.1151 & \$0.1179 \\
Gemini 3 Flash & 0.56 & 82.5 & 16.1 & \$0.0125 & \$0.0151 \\
Gemini 3.1 Pro & 0.55 & 146.5 & 10.5 & \$0.0380 & \$0.0400 \\
GPT-4o Mini & 0.48 & 65.0 & 20.0 & \$0.0041 & \$0.0070 \\
Claude Opus 4.7 & 0.47 & 103.5 & 17.3 & \$0.1291 & \$0.1321 \\
DeepSeek V4 Flash & 0.47 & 160.5 & 18.4 & \$0.0035 & \$0.0064 \\
DeepSeek V4 Pro & 0.45 & 264.0 & 17.1 & \$0.0103 & \$0.0130 \\
Mistral Small 24B & 0.45 & 119.5 & 18.7 & \$0.0039 & \$0.0069 \\
Claude Haiku 4.5 & 0.45 & 98.5 & 18.7 & \$0.0271 & \$0.0302 \\
Ministral 3 14B & 0.43 & 100.0 & 18.8 & \$0.0049 & \$0.0079 \\
Qwen 3.5 35B & 0.41 & 208.5 & 17.3 & \$0.0042 & \$0.0070 \\
Qwen 3.5 9B & 0.34 & 92.5 & 19.4 & \$0.0026 & \$0.0056 \\
GPT-5.4 Mini & 0.34 & 80.0 & 19.1 & \$0.0209 & \$0.0239 \\
\midrule
Assistant overhead & --- & --- & --- & --- & \$0.0028 \\
\multicolumn{6}{l}{\footnotesize\textit{(always GPT-4o Mini, charged every episode regardless of instructor)}} \\
\bottomrule
\end{tabular}
\end{table*}

\begin{figure}[!htbp]
  \centering
  \includegraphics[width=\columnwidth]{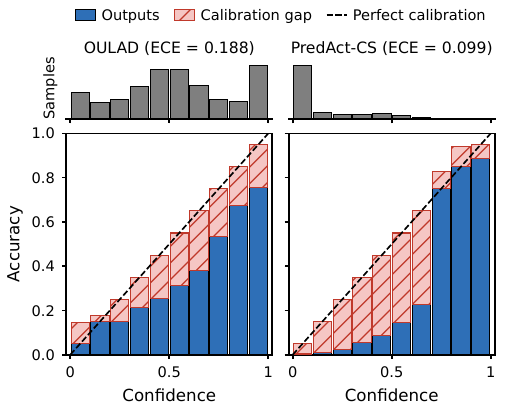}
  \caption{Reliability diagrams of the k-NN grade predictor on OULAD (left) and \predactcs (right). The diagonal shows perfect calibration; blue bars are observed accuracy within each confidence bin, and the red hatched region is the calibration gap. Top histograms show the distribution of predictions across confidence bins. Both predictors are systematically overconfident, with OULAD exhibiting a larger expected calibration error (ECE $= 0.188$) than \predactcs (ECE $= 0.099$).}
  \label{fig:reliability}
\end{figure}
\begin{figure}[!htbp]
  \centering
  \includegraphics[width=\columnwidth]{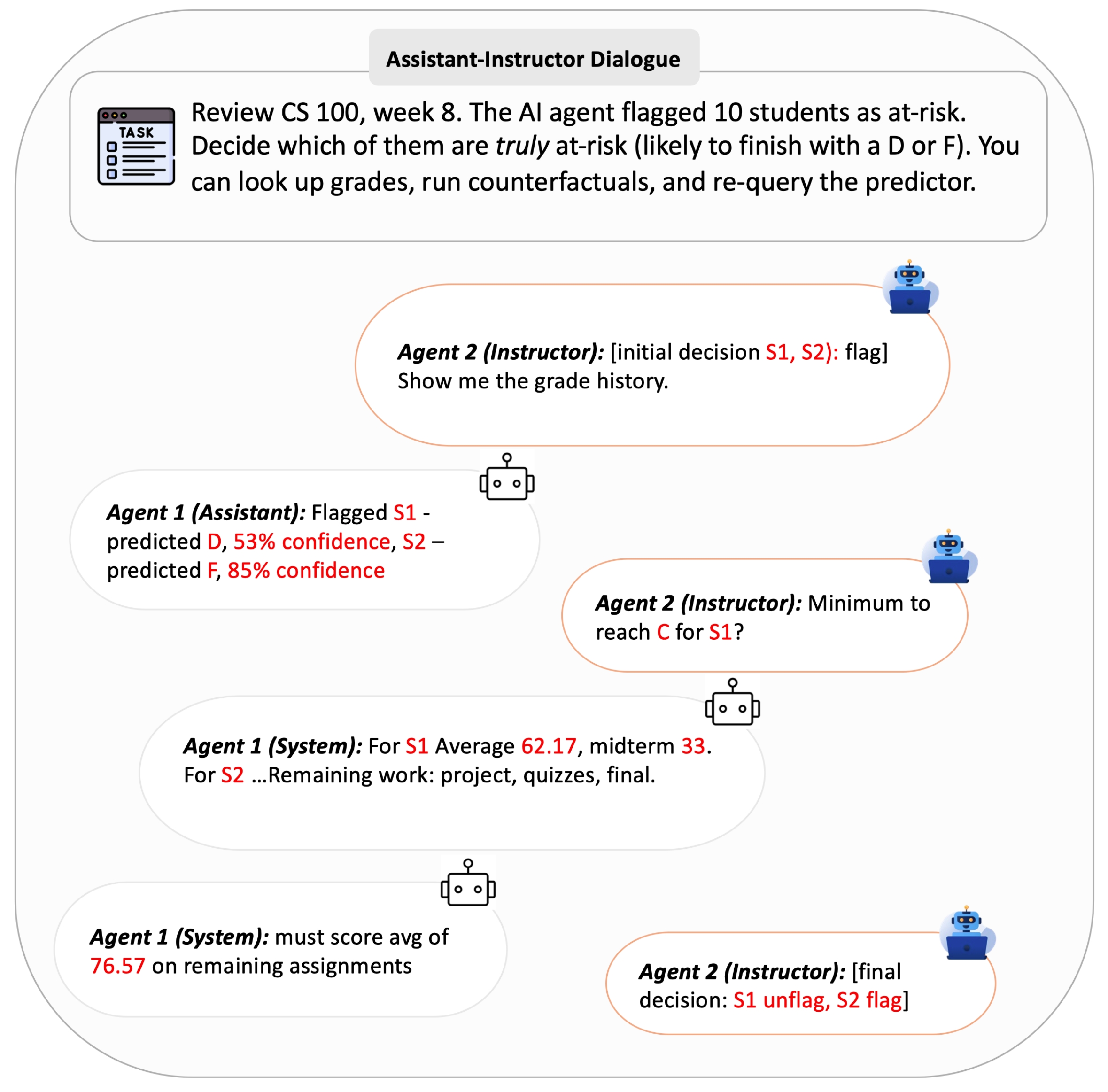}
  \caption{Example \predact episode. The instructor agent receives a 
  task description, sees flagged students with predicted grades and 
  confidence scores, records an initial decision, then queries the 
  assistant agent for grade histories and counterfactuals before 
  committing to a final decision. Red text indicates values returned 
  by tools.}
  \label{fig:my-figure}
\end{figure}

\begin{figure*}[t]
  \centering
  \includegraphics[width=\textwidth]{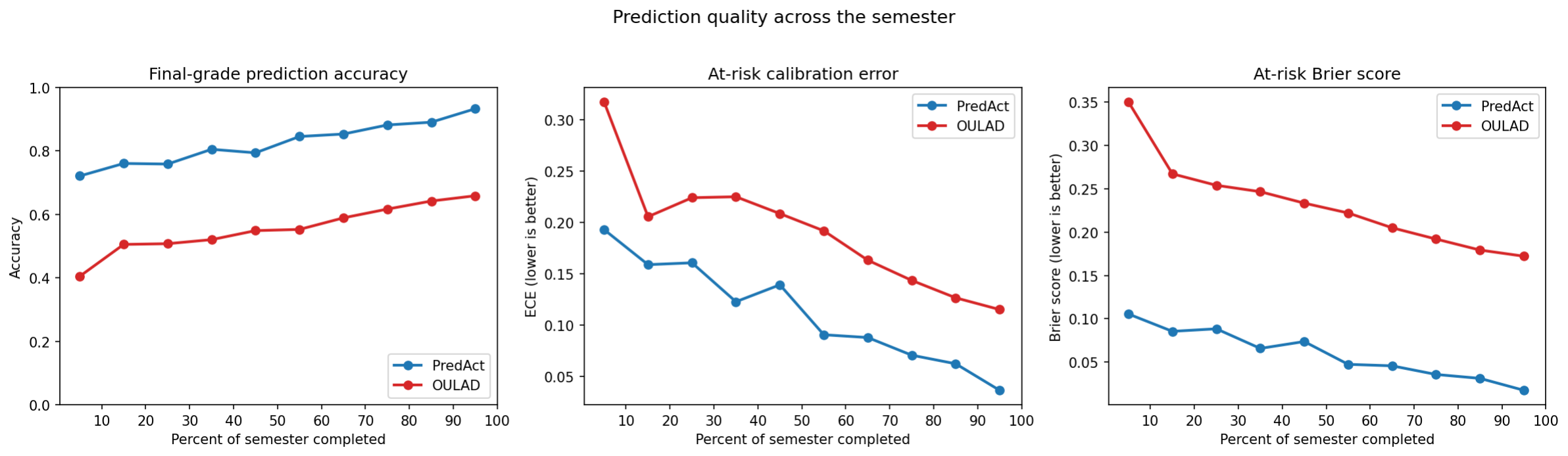}
\caption{Tool quality across the semester. Accuracy, ECE, and Brier score for
the k-NN grade predictor on \predactcs and OULAD, showing that tool outputs are
realistically noisy at every point in the term.}
\label{fig:exp1_over_time}
  \label{fig:dialogue}
\end{figure*}

\subsection{Assistant Ablation}
\label{app:assistant_ablation}

Table~\ref{tab:assistant_ablation_full} reports the full means for Section~\ref{sec:ablations}. Each assistant has 90 episodes. The GPT-4o Mini baseline comes from the main benchmark, and each replacement assistant has 90 additional episodes. Student samples, sampling and noise seeds, and the tool's binary flags match within each comparison. Sampling seeds do not fix LLM generation, so initial decisions can differ.

We average each of the nine instructor and accuracy combinations equally. Within a combination, each metric excludes only its unavailable values. The saved evaluator omits initial F1 when no students are initially accepted, even when those decisions are recorded. Initial F1 uses 83, 83, and 85 episodes for GPT-4o Mini, Claude Opus 4.7, and DeepSeek V4 Pro, respectively, while final F1 uses all 90 for each assistant. $\Delta F_1$ subtracts these means rather than averaging paired episode changes. RAIR uses 50, 48, and 52 episodes, and RSR uses 87 for each assistant.

\begin{table}[!htbp]
\centering
\footnotesize
\setlength{\tabcolsep}{3.5pt}
\caption{Assistant ablation means over nine instructor and accuracy combinations. F1, RAIR, and RSR use the 0 to 1 scale. $\Delta F_1$ is computed before rounding.}
\label{tab:assistant_ablation_full}
\begin{tabular}{lrrrrr}
\toprule
Assistant & \makecell{Initial\\F1} & \makecell{Final\\F1} & $\Delta F_1$ & RAIR & RSR \\
\midrule
GPT-4o Mini & 0.590 & 0.520 & $-0.070$ & 0.137 & 0.970 \\
Claude Opus 4.7 & 0.571 & 0.533 & $-0.038$ & 0.235 & 0.971 \\
DeepSeek V4 Pro & 0.574 & 0.465 & $-0.110$ & 0.281 & 0.911 \\
\bottomrule
\end{tabular}
\end{table}

The two sided paired $t$ tests compare $\Delta F_1$ across nine matching instructor and accuracy combinations. All three tests have $p>0.05$. This small comparison does not establish equivalence.

\subsection{Structured Noise Ablation}
\label{app:structured_noise}

The comparison uses the same instructors, accuracy levels, and runs as the assistant ablation, with GPT-4o Mini as assistant throughout. Uniform errors come from the same 90 original episodes, and borderline errors use 90 additional episodes. The latter concentrate flips near the C grade boundary separating at risk grades D and F from grades C and above. Student samples and sampling and noise seeds are matched, and calibration preserves target letter grade accuracy. The sets of flagged students can differ.

For Table~\ref{tab:structured_noise_full}, we first exclude episodes whose initial F1 is omitted by the saved evaluator, leaving 83 uniform and 87 borderline episodes. RAIR and RSR then exclude their own empty denominators. RAIR uses 46 and 32 episodes, and RSR uses 81 and 67, respectively. We average the nine instructor and accuracy combinations equally. This differs from the metric specific filtering in Table~\ref{tab:assistant_ablation_full} and explains the different baseline means.

\begin{table}[!htbp]
\centering
\footnotesize
\setlength{\tabcolsep}{3.5pt}
\caption{Structured noise ablation means. F1, RAIR, and RSR use the 0 to 1 scale. Initial and final F1 use the same eligible episodes within each condition, and $\Delta F_1$ is computed before rounding.}
\label{tab:structured_noise_full}
\begin{tabular}{lrrrrr}
\toprule
\makecell[l]{Error\\placement} & \makecell{Initial\\F1} & \makecell{Final\\F1} & $\Delta F_1$ & RAIR & RSR \\
\midrule
Uniform & 0.590 & 0.547 & $-0.043$ & 0.133 & 0.974 \\
Borderline & 0.486 & 0.500 & $+0.015$ & 0.365 & 0.996 \\
\bottomrule
\end{tabular}
\end{table}

At 80\% target accuracy, initial F1 is 0.850 under uniform errors and 0.595 under borderline errors, while final F1 is 0.733 and 0.555. The final score gap therefore cannot be attributed to dialogue alone. Across all three accuracy levels, the positive mean change under borderline errors does not reproduce the uniform condition's mean decline. Its RAIR of 0.365 indicates incomplete correction of initial mistakes when the tool is right, while RSR of 0.996 indicates correct resistance when the tool is wrong.

\section{Expert Evaluation of Dialogue Quality}
\label{app:manual_eval}

To complement the automated behavior metrics in the main paper, two expert raters manually scored a stratified sample of human and LLM dialogues on two dialogue-quality dimensions. Each dialogue was scored on a 0--2 scale: verification quality captures whether the rater actively checked the agent's call before deciding, and question quality captures whether the rater's questions were specific, targeted, and built on prior tool answers. Five dialogues were double-coded as a calibration check; raters agreed exactly on 4/5 verification scores and 3/5 question scores. Significance is reported using the Mann-Whitney U test, appropriate for the ordinal 0--2 scale.

Table~\ref{tab:manual_eval} reports the results. Verification quality is statistically indistinguishable between humans and LLMs (both 1.42, $p=0.826$), confirming that mechanical thoroughness is not the differentiator. Question quality, however, is significantly higher for humans (1.54 vs 1.21, $p=0.026$), supporting the qualitative finding from the main paper: humans ask better, more targeted, more adaptive questions, even when they ask fewer of them.
\begin{table}[!htbp]
\centering
\footnotesize
\setlength{\tabcolsep}{4pt}
\renewcommand{\arraystretch}{1.1}
\caption{Manual evaluation of dialogue quality. Two expert raters scored sampled human and LLM dialogues on a 0--2 scale across two dimensions: \textbf{verification quality} (whether the rater checked the agent's call) and \textbf{question quality} (whether the questions were specific, built on prior answers, and included follow-ups). $p$-values are from Mann-Whitney U tests comparing humans vs LLMs.}
\label{tab:manual_eval}
\begin{tabular*}{\linewidth}{@{\extracolsep{\fill}} l ccc @{}}
\toprule
\textbf{Dimension} & \makecell{\textbf{Human}\\($n=26$)} & \makecell{\textbf{LLM}\\($n=24$)} & \textbf{$p$-value} \\
\midrule
Verification quality & 1.42 & 1.42 & 0.826~(n.s.) \\
Question quality     & \textbf{1.54} & 1.21 & \textbf{0.026} \\
\bottomrule
\end{tabular*}
\end{table}

\section{F1 Results}
\label{app:f1}
\begin{figure*}[t]
  \centering
  \includegraphics[width=\textwidth]{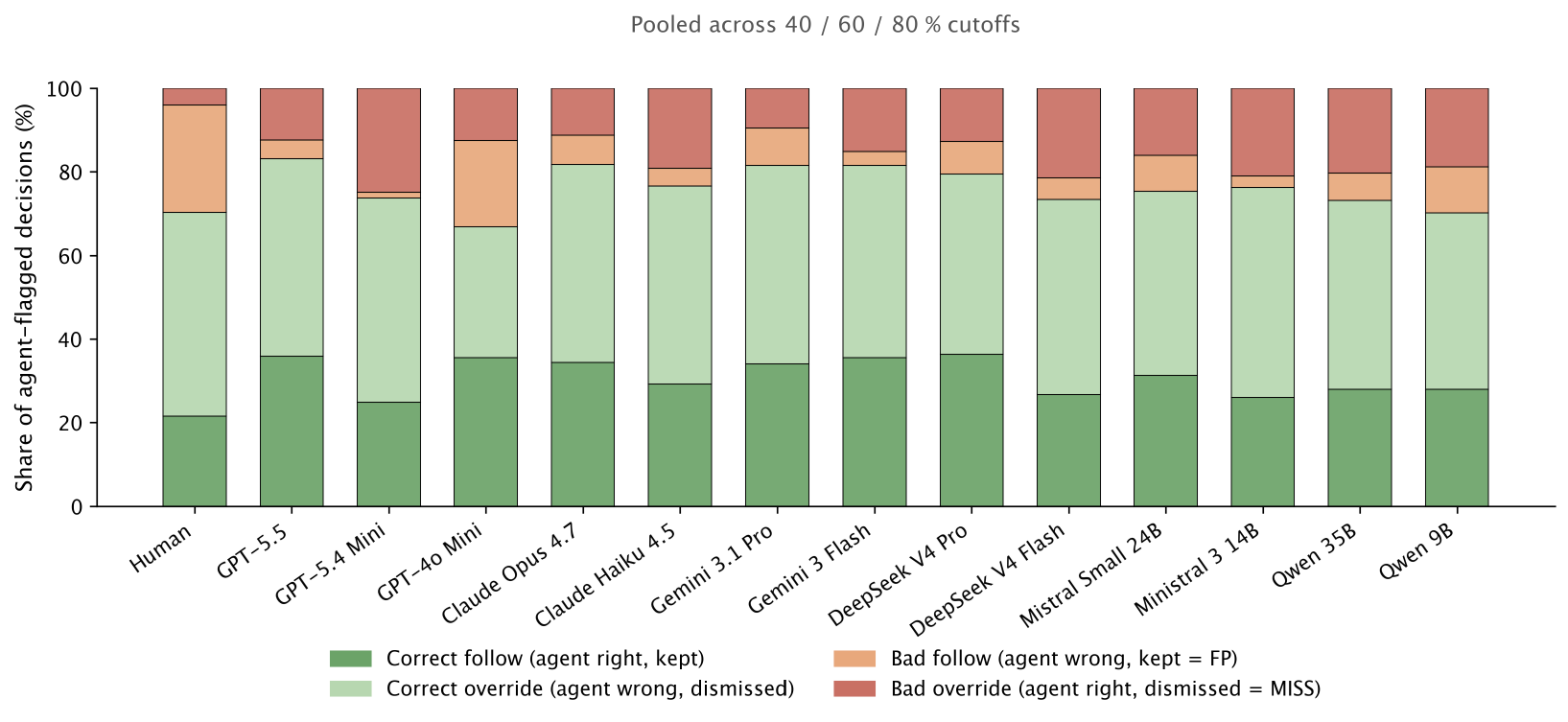}
  \caption{Override behavior on agent-flagged students for humans (n=13, GPT-4o Mini assistant) vs. 13 LLM instructors, all paired with the same calibrated synthetic tool and GPT-4o Mini assistant. Bars show the share of decisions in four ground-truth-aware buckets: correct follow / correct override / bad follow / bad override. Pooled across 40 / 60 / 80 \% target tool accuracies. \predactcs only.}
  \label{fig:override}
\end{figure*}

\begin{figure*}[t]
\centering
\includegraphics[width=0.75\linewidth,trim={79.5bp 0 -79.5bp 0}]{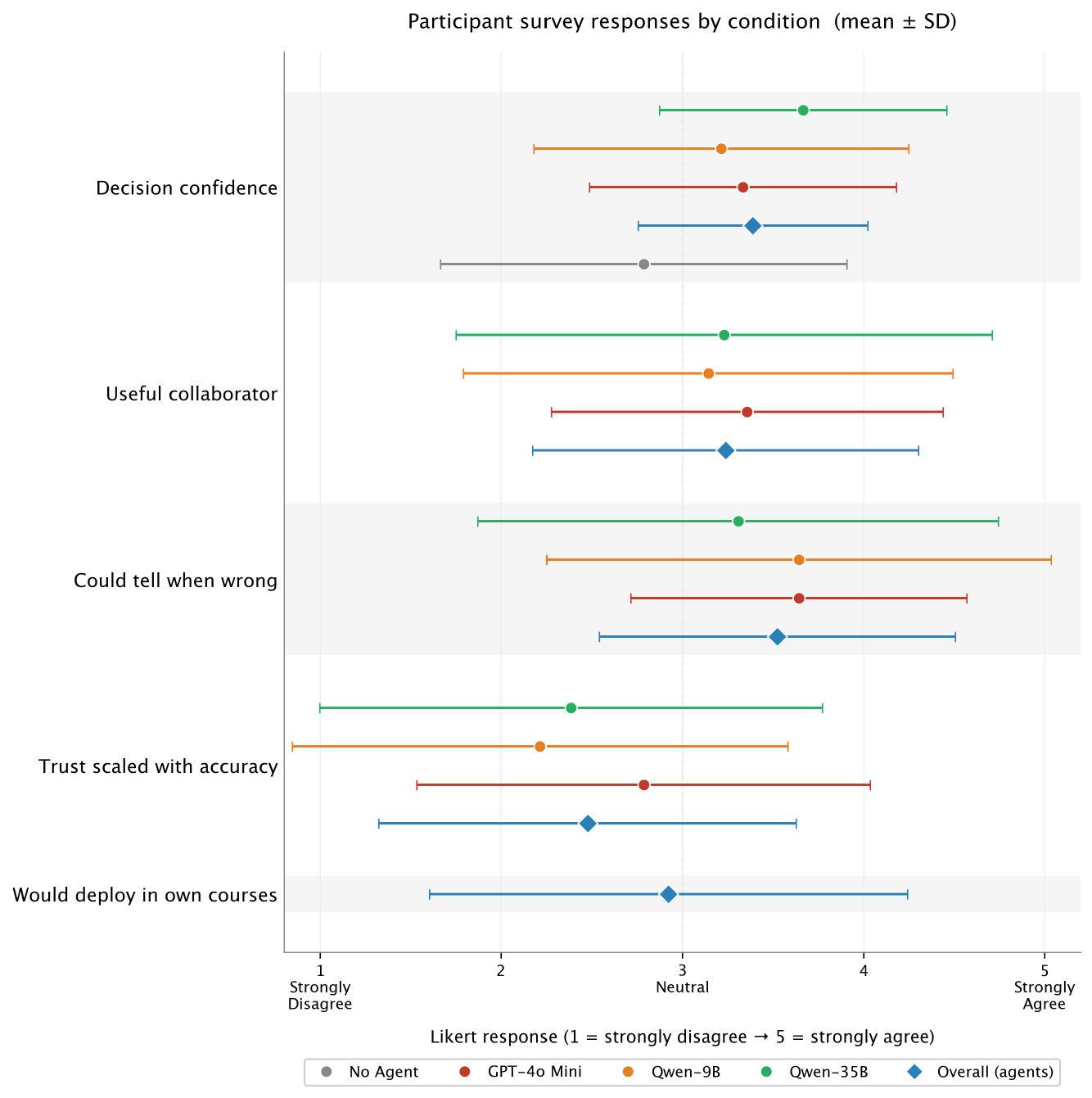}
\caption{Post-study Likert survey responses (mean $\pm$ SD, $n=13$) across five questions and four conditions. \textit{Decision confidence} is reported for all conditions including the no-agent baseline; the remaining four questions apply only to agent conditions. \textit{No Agent} (gray) serves as a reference baseline. Agent conditions are GPT-4o Mini, Qwen-9B, and Qwen-35B; \textit{Overall (agents)} (diamond) pools responses across all three. The dashed vertical line marks the neutral midpoint (3). Scale: 1 = Strongly Disagree, 5 = Strongly Agree.}
\label{fig:likert_survey}
\end{figure*}

Tables~\ref{tab:main_results} and~\ref{tab:trust_calibration} report F1 results for the LLM instructors, Tables~\ref{tab:human_results} and~\ref{tab:human_results_f1} report them for the human study, and Figures~\ref{fig:override} and~\ref{fig:likert_survey} show override behavior and survey responses.

\label{app:per_model_metrics}

\begin{table}[!htbp]
\centering
\footnotesize
\setlength{\tabcolsep}{3pt}
\renewcommand{\arraystretch}{1.1}
\caption{Per-model breakdown of initial F1, final F1, RAIR, and RSR, 
averaged across all 10 cells (5 target accuracies $\times$ 2 datasets, 
10 runs per cell). $\Delta$F1 is the change from initial to final 
($d_i^{\text{init}} \to d_i^{\text{final}}$). All values $\times 100$. 
Models sorted by final F1.}
\label{tab:trust_calibration}
\begin{tabular*}{\linewidth}{@{\extracolsep{\fill}} l ccccc @{}}
\toprule
\textbf{Model} & \textbf{F1$_{\text{init}}$} & \textbf{F1$_{\text{final}}$} & \textbf{$\Delta$F1} & \textbf{RAIR} & \textbf{RSR} \\
\midrule
GPT-5.5            & 56.7 & 57.2 & $+0.5$  & 49.4 & 90.5 \\
Gemini 3 Flash     & 55.6 & 56.0 & $+0.4$  & 16.7 & 100.0 \\
Gemini 3.1 Pro     & 53.2 & 55.1 & $+1.9$  & 32.7 & 95.4 \\
Claude Opus 4.7    & 51.4 & 47.2 & $-4.2$  &  7.1 & 98.5 \\
GPT-4o Mini        & 50.7 & 47.8 & $-2.9$  &  0.0 & 99.7 \\
DeepSeek V4 Flash  & 54.3 & 47.0 & $-7.3$  &  9.6 & 99.4 \\
Mistral Small 24B  & 47.7 & 44.9 & $-2.8$  &  5.0 & 100.0 \\
DeepSeek V4 Pro    & 51.0 & 44.9 & $-6.1$  &  8.1 & 99.7 \\
Claude Haiku 4.5   & 49.5 & 44.7 & $-4.8$  & 13.7 & 98.7 \\
Ministral 3 14B    & 49.5 & 42.0 & $-7.5$  &  4.9 & 98.5 \\
Qwen 3.5 35B           & 47.6 & 41.8 & $-5.8$  & 15.0 & 98.0 \\
Qwen 3.5 9B            & 42.4 & 33.9 & $-8.5$  &  4.3 & 98.7 \\
GPT-5.4 Mini       & 44.6 & 33.6 & $-11.0$ &  5.6 & 99.4 \\
\midrule
\textbf{Mean}      & 50.3 & 45.9 & $-4.5$  & 13.2 & 98.2 \\
\bottomrule
\end{tabular*}
\end{table}

\begin{table}[!htbp]
\centering
\footnotesize
\setlength{\tabcolsep}{4pt}
\renewcommand{\arraystretch}{1.1}
\caption{Human study: F1 ($\times 100$) of instructors and teaching assistants paired with three LLM assistants at target tool accuracies (40\%--80\%). \textbf{Init} is the participant's at-risk classification F1 before agent dialogue; \textbf{Final} is after dialogue. \textbf{Overall} is the per-agent mean F1 across cutoffs. Best per column within each block in \textbf{bold}. Each cell aggregates 13 participants. The no-agent baseline (humans deciding without any LLM agent) achieves Final F1 = 89.0.}
\label{tab:human_results}
\begin{tabular*}{\linewidth}{@{\extracolsep{\fill}} l l c ccc @{}}
\toprule
\textbf{Agent} & \textbf{Stage} & \textbf{Overall} & \textbf{40\%} & \textbf{60\%} & \textbf{80\%} \\
\midrule
\rowcolor{gray!20} \multicolumn{6}{c}{\textit{Initial decision (pre-dialogue)}} \\
\midrule
GPT-4o Mini & Init  & 49.6 & \textbf{34.8} & 59.0 & 54.9 \\
Qwen-9B     & Init  & 50.8 & 30.8 & \textbf{67.7} & 53.8 \\
Qwen-35B    & Init  & \textbf{52.8} & 33.6 & 66.2 & \textbf{58.7} \\
\midrule
\rowcolor{gray!20} \multicolumn{6}{c}{\textit{Final decision (post-dialogue)}} \\
\midrule
GPT-4o Mini & Final & 55.3 & 27.7 & 68.7 & \textbf{69.5} \\
Qwen-9B     & Final & 55.8 & 31.6 & 72.6 & 63.3 \\
Qwen-35B    & Final & \textbf{56.3} & \textbf{32.6} & \textbf{72.8} & 63.6 \\
\bottomrule
\end{tabular*}
\end{table}

\begin{table*}[t]
\centering
\footnotesize
\setlength{\tabcolsep}{8pt}
\renewcommand{\arraystretch}{1.1}
\caption{F1 ($\times 100$, mean\,$\pm$\,std) of LLM instructors on \predactcs and OULAD across target tool accuracies (40\%--80\%). \textbf{Overall} is the per-dataset mean F1 across cutoffs. Best per column within each block (per dataset) in \textbf{bold}. Rows sorted within each block by combined Overall across both datasets. Each cell aggregates 10 episodes (30 students, 5 forced at-risk).}
\label{tab:main_results}
\begin{tabular*}{\textwidth}{@{\extracolsep{\fill}} l l c ccccc @{}}
\toprule
\textbf{Model} & \textbf{Dataset} & \textbf{Overall} & \textbf{40\%} & \textbf{50\%} & \textbf{60\%} & \textbf{70\%} & \textbf{80\%} \\
\midrule
\rowcolor{gray!20} \multicolumn{8}{c}{\textit{Closed-source Models}} \\
\midrule
\multirow{2}{*}{GPT-5.5} & \predactcs & 59.2 & 34.8\,\tiny$\pm$32.2 & \textbf{74.7}\,\tiny$\pm$13.3 & \textbf{57.8}\,\tiny$\pm$10.4 & 46.4\,\tiny$\pm$33.9 & \textbf{82.6}\,\tiny$\pm$13.5 \\
& OULAD & \textbf{55.1} & \textbf{44.5}\,\tiny$\pm$25.4 & \textbf{43.5}\,\tiny$\pm$22.0 & \textbf{61.3}\,\tiny$\pm$25.3 & \textbf{75.4}\,\tiny$\pm$14.9 & 51.0\,\tiny$\pm$28.9 \\
\multirow{2}{*}{Gemini 3 Flash} & \predactcs & \textbf{63.4} & \textbf{55.8}\,\tiny$\pm$16.4 & 70.3\,\tiny$\pm$13.7 & 50.6\,\tiny$\pm$14.3 & \textbf{59.8}\,\tiny$\pm$21.5 & 80.6\,\tiny$\pm$15.5 \\
& OULAD & 48.6 & 29.5\,\tiny$\pm$27.0 & 30.2\,\tiny$\pm$18.2 & 57.7\,\tiny$\pm$15.2 & 61.9\,\tiny$\pm$18.2 & 63.5\,\tiny$\pm$15.9 \\
\multirow{2}{*}{Gemini 3.1 Pro} & \predactcs & 57.0 & 38.9\,\tiny$\pm$28.8 & 62.2\,\tiny$\pm$18.2 & 44.7\,\tiny$\pm$12.0 & 58.9\,\tiny$\pm$11.0 & 80.5\,\tiny$\pm$15.7 \\
& OULAD & 52.3 & 43.7\,\tiny$\pm$21.2 & 36.2\,\tiny$\pm$26.0 & 55.6\,\tiny$\pm$25.8 & 56.2\,\tiny$\pm$23.8 & \textbf{70.1}\,\tiny$\pm$14.1 \\
\multirow{2}{*}{Claude Opus 4.7} & \predactcs & 55.4 & 31.1\,\tiny$\pm$26.2 & 68.1\,\tiny$\pm$13.8 & 47.4\,\tiny$\pm$30.0 & 55.3\,\tiny$\pm$28.0 & 75.3\,\tiny$\pm$20.5 \\
& OULAD & 38.5 & 14.8\,\tiny$\pm$20.6 & 38.3\,\tiny$\pm$22.4 & 39.4\,\tiny$\pm$28.9 & 47.9\,\tiny$\pm$23.0 & 52.3\,\tiny$\pm$24.2 \\
\multirow{2}{*}{GPT-4o Mini} & \predactcs & 54.2 & 42.4\,\tiny$\pm$16.9 & 55.5\,\tiny$\pm$14.1 & 47.9\,\tiny$\pm$13.1 & 51.8\,\tiny$\pm$16.6 & 73.6\,\tiny$\pm$11.4 \\
& OULAD & 41.5 & 28.6\,\tiny$\pm$20.2 & 28.8\,\tiny$\pm$20.6 & 43.3\,\tiny$\pm$11.9 & 56.7\,\tiny$\pm$16.3 & 49.9\,\tiny$\pm$13.2 \\
\multirow{2}{*}{Claude Haiku 4.5} & \predactcs & 51.9 & 31.9\,\tiny$\pm$31.2 & 44.7\,\tiny$\pm$22.5 & 55.5\,\tiny$\pm$18.5 & 59.5\,\tiny$\pm$16.2 & 67.9\,\tiny$\pm$18.5 \\
& OULAD & 37.5 & 26.4\,\tiny$\pm$21.4 & 29.1\,\tiny$\pm$17.5 & 28.6\,\tiny$\pm$23.1 & 55.0\,\tiny$\pm$18.2 & 48.3\,\tiny$\pm$22.8 \\
\multirow{2}{*}{GPT-5.4 Mini} & \predactcs & 44.4 & 40.4\,\tiny$\pm$25.2 & 54.4\,\tiny$\pm$14.1 & 43.5\,\tiny$\pm$15.4 & 23.3\,\tiny$\pm$27.4 & 60.4\,\tiny$\pm$19.0 \\
& OULAD & 23.2 & 12.9\,\tiny$\pm$16.7 & 28.6\,\tiny$\pm$16.9 & 32.9\,\tiny$\pm$24.9 & 26.5\,\tiny$\pm$26.4 & 15.2\,\tiny$\pm$16.2 \\
\midrule
\rowcolor{gray!20} \multicolumn{8}{c}{\textit{Open-source Models}} \\
\midrule
\multirow{2}{*}{DeepSeek V4 Flash} & \predactcs & 52.4 & 28.8\,\tiny$\pm$18.1 & 62.2\,\tiny$\pm$19.3 & 49.7\,\tiny$\pm$22.5 & 55.4\,\tiny$\pm$20.2 & 65.7\,\tiny$\pm$28.7 \\
& OULAD & \textbf{41.3} & \textbf{29.2}\,\tiny$\pm$23.5 & \textbf{32.5}\,\tiny$\pm$19.9 & 35.0\,\tiny$\pm$16.9 & \textbf{52.3}\,\tiny$\pm$27.8 & \textbf{57.6}\,\tiny$\pm$33.4 \\
\multirow{2}{*}{DeepSeek V4 Pro} & \predactcs & \textbf{60.7} & 42.6\,\tiny$\pm$22.6 & \textbf{68.5}\,\tiny$\pm$19.1 & 48.2\,\tiny$\pm$20.6 & 60.3\,\tiny$\pm$19.3 & \textbf{83.8}\,\tiny$\pm$13.2 \\
& OULAD & 29.9 & 27.8\,\tiny$\pm$21.8 & 23.5\,\tiny$\pm$20.0 & 32.4\,\tiny$\pm$19.3 & 33.1\,\tiny$\pm$27.4 & 33.0\,\tiny$\pm$26.9 \\
\multirow{2}{*}{Mistral Small 24B} & \predactcs & 57.0 & \textbf{42.8}\,\tiny$\pm$20.3 & 58.5\,\tiny$\pm$16.7 & 51.6\,\tiny$\pm$14.5 & \textbf{62.2}\,\tiny$\pm$21.6 & 70.0\,\tiny$\pm$16.7 \\
& OULAD & 32.3 & 13.0\,\tiny$\pm$17.4 & 15.6\,\tiny$\pm$18.0 & 33.7\,\tiny$\pm$22.7 & 52.3\,\tiny$\pm$24.2 & 47.1\,\tiny$\pm$26.5 \\
\multirow{2}{*}{Ministral 3 14B} & \predactcs & 49.2 & 34.8\,\tiny$\pm$15.7 & 49.1\,\tiny$\pm$28.1 & 47.8\,\tiny$\pm$28.1 & 44.8\,\tiny$\pm$18.4 & 69.8\,\tiny$\pm$16.4 \\
& OULAD & 36.1 & 26.3\,\tiny$\pm$21.3 & 22.5\,\tiny$\pm$17.9 & \textbf{39.5}\,\tiny$\pm$18.2 & 48.8\,\tiny$\pm$19.3 & 43.5\,\tiny$\pm$25.4 \\
\multirow{2}{*}{Qwen 3.5 35B} & \predactcs & 49.8 & 27.7\,\tiny$\pm$23.1 & 61.2\,\tiny$\pm$18.1 & 40.6\,\tiny$\pm$21.2 & 50.1\,\tiny$\pm$17.7 & 69.5\,\tiny$\pm$24.2 \\
& OULAD & 31.5 & 22.6\,\tiny$\pm$22.6 & 21.1\,\tiny$\pm$20.0 & 29.5\,\tiny$\pm$23.4 & 45.8\,\tiny$\pm$15.0 & 38.7\,\tiny$\pm$13.2 \\
\multirow{2}{*}{Qwen 3.5 9B} & \predactcs & 43.9 & 22.2\,\tiny$\pm$24.8 & 37.2\,\tiny$\pm$26.4 & \textbf{54.7}\,\tiny$\pm$23.2 & 38.8\,\tiny$\pm$29.8 & 66.8\,\tiny$\pm$22.2 \\
& OULAD & 24.6 & 12.4\,\tiny$\pm$14.1 & 22.5\,\tiny$\pm$14.1 & 24.7\,\tiny$\pm$20.3 & 50.7\,\tiny$\pm$24.6 & 12.5\,\tiny$\pm$22.7 \\
\bottomrule
\end{tabular*}
\end{table*}
\begin{table*}[t]
\centering
\footnotesize
\caption{F1 ($\times 100$, mean$\,\pm\,$std) of 13 participants paired with three LLM assistants on \predactcs across target tool accuracies (40\%--80\%).}
\label{tab:human_results_f1}
\begin{tabular*}{\textwidth}{@{\extracolsep{\fill}} l l c ccc @{}}
\toprule
\textbf{LLM Assistant} & \textbf{Dataset} & \textbf{Overall} & \textbf{40\%} & \textbf{60\%} & \textbf{80\%} \\
\midrule
No agent (baseline) & \predactcs & 89.0\,\tiny$\pm$14.1 & -- & -- & -- \\
\midrule
GPT-4o Mini & \predactcs & 55.3\,\tiny$\pm$35.4 & 27.7\,\tiny$\pm$36.4 & 68.7\,\tiny$\pm$32.4 & \textbf{69.5}\,\tiny$\pm$15.5 \\
Qwen-35B    & \predactcs & \textbf{56.3}\,\tiny$\pm$35.0 & \textbf{32.6}\,\tiny$\pm$36.0 & \textbf{72.8}\,\tiny$\pm$27.2 & 63.6\,\tiny$\pm$27.3 \\
Qwen-9B     & \predactcs & 55.8\,\tiny$\pm$30.3 & 31.6\,\tiny$\pm$36.0 & 72.6\,\tiny$\pm$18.9 & 63.3\,\tiny$\pm$13.4 \\
\bottomrule
\end{tabular*}
\end{table*}

\section{Q-Type Definitions}
\label{app:q_dist}
Table~\ref{tab:qtype_definitions} defines the six question categories used to classify rater messages.
\begin{table*}[t]
\centering
\footnotesize
\setlength{\tabcolsep}{8pt}
\renewcommand{\arraystretch}{1.2}
\caption{Definitions of the six question categories used to classify rater messages in Table~\ref{tab:behavior_profile} and the Q-type distribution analysis.}
\label{tab:qtype_definitions}
\begin{tabular*}{\linewidth}{@{\extracolsep{\fill}} l p{0.72\linewidth} @{}}
\toprule
\textbf{Category} & \textbf{Description} \\
\midrule
\texttt{grade\_lookup}     & Asking for a specific student's current grades or assignment scores. \\
\texttt{min\_score}        & Asking what score a student needs on remaining work to reach a target grade. \\
\texttt{counterfactual}    & ``What if'' scenarios (e.g., what if the student scores $X$ on the final?). \\
\texttt{prediction\_query} & Asking for the model's predicted final grade or pass/fail outcome. \\
\texttt{class\_comparison} & Comparing a student against the class average or other students. \\
\texttt{other}             & Anything not covered above (e.g., general course information, listing flagged students). \\
\bottomrule
\end{tabular*}
\end{table*}

\section{Human study interface}
\label{app:human-study-interface}

Here we present the interface for the human study along with the instructions.

\begin{figure*}[t]
    \centering
    \includegraphics[width=1\linewidth]{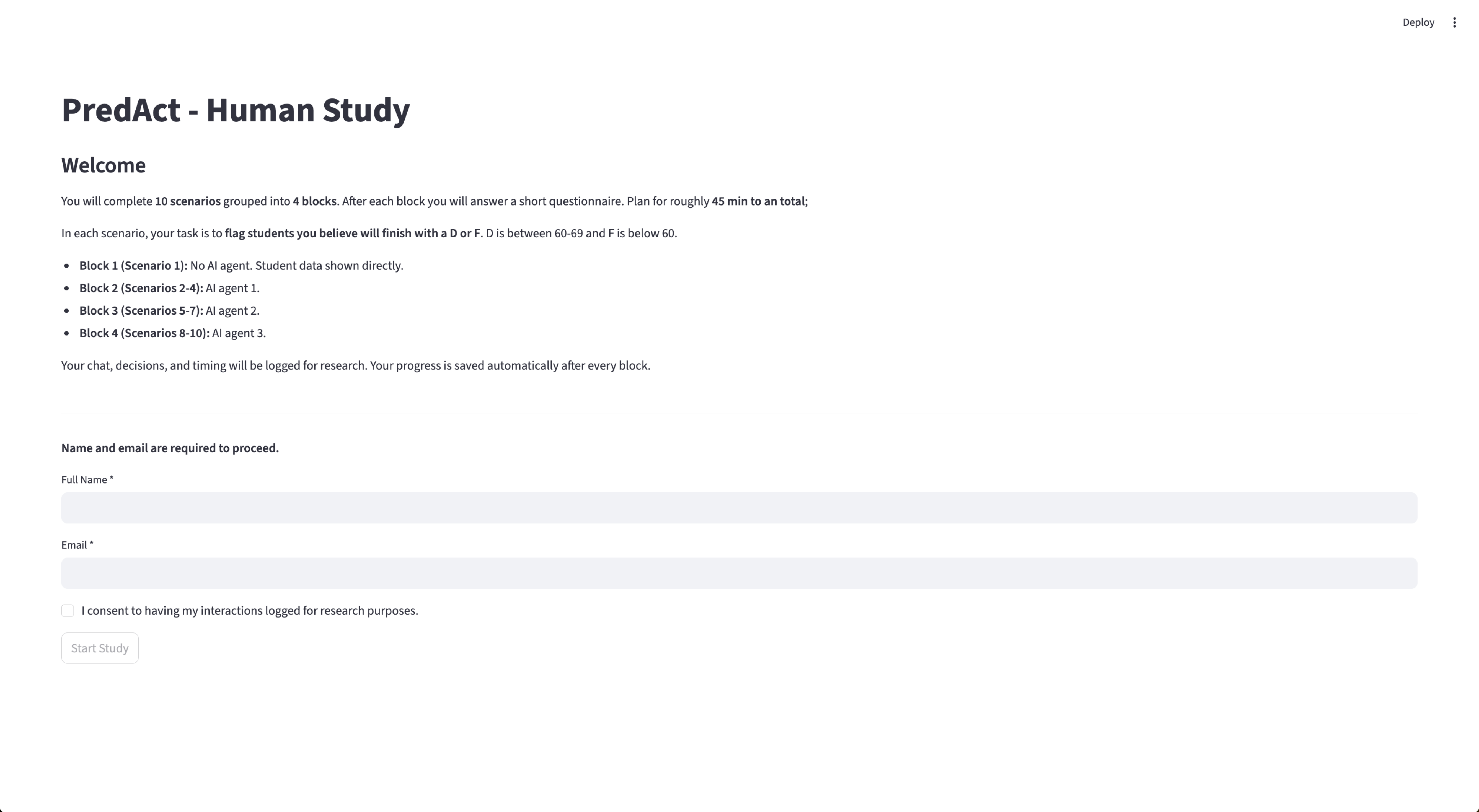}
    \caption{Welcome page of the human study.}
    \label{fig:human study 1}
\end{figure*}

\begin{figure*}[t]
    \centering
    \includegraphics[width=1\linewidth]{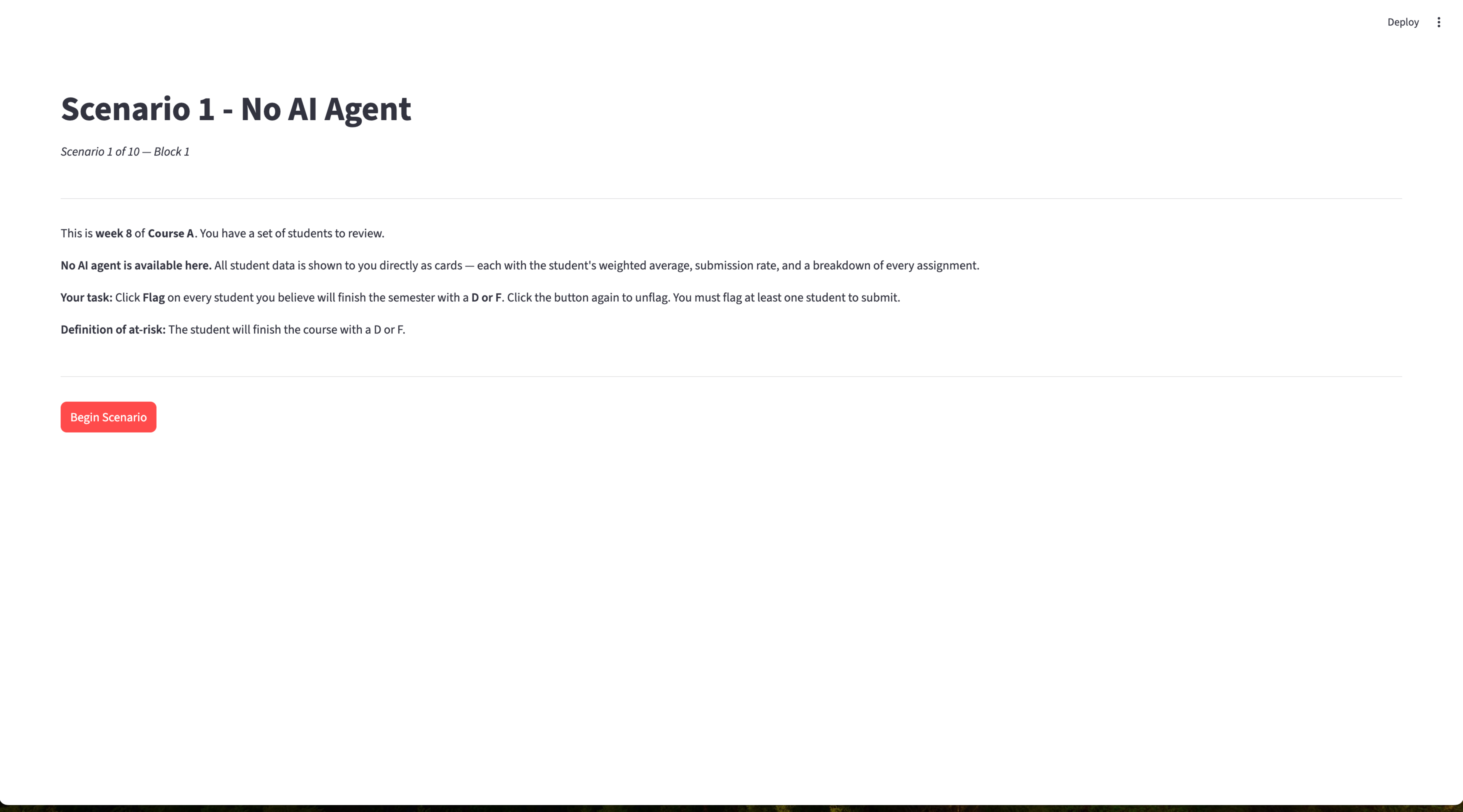}
    \caption{Instruction page for beginning the no-agent scenario.}
    \label{fig:human study 2}
\end{figure*}

\begin{figure*}[t]
    \centering
    \includegraphics[width=1\linewidth]{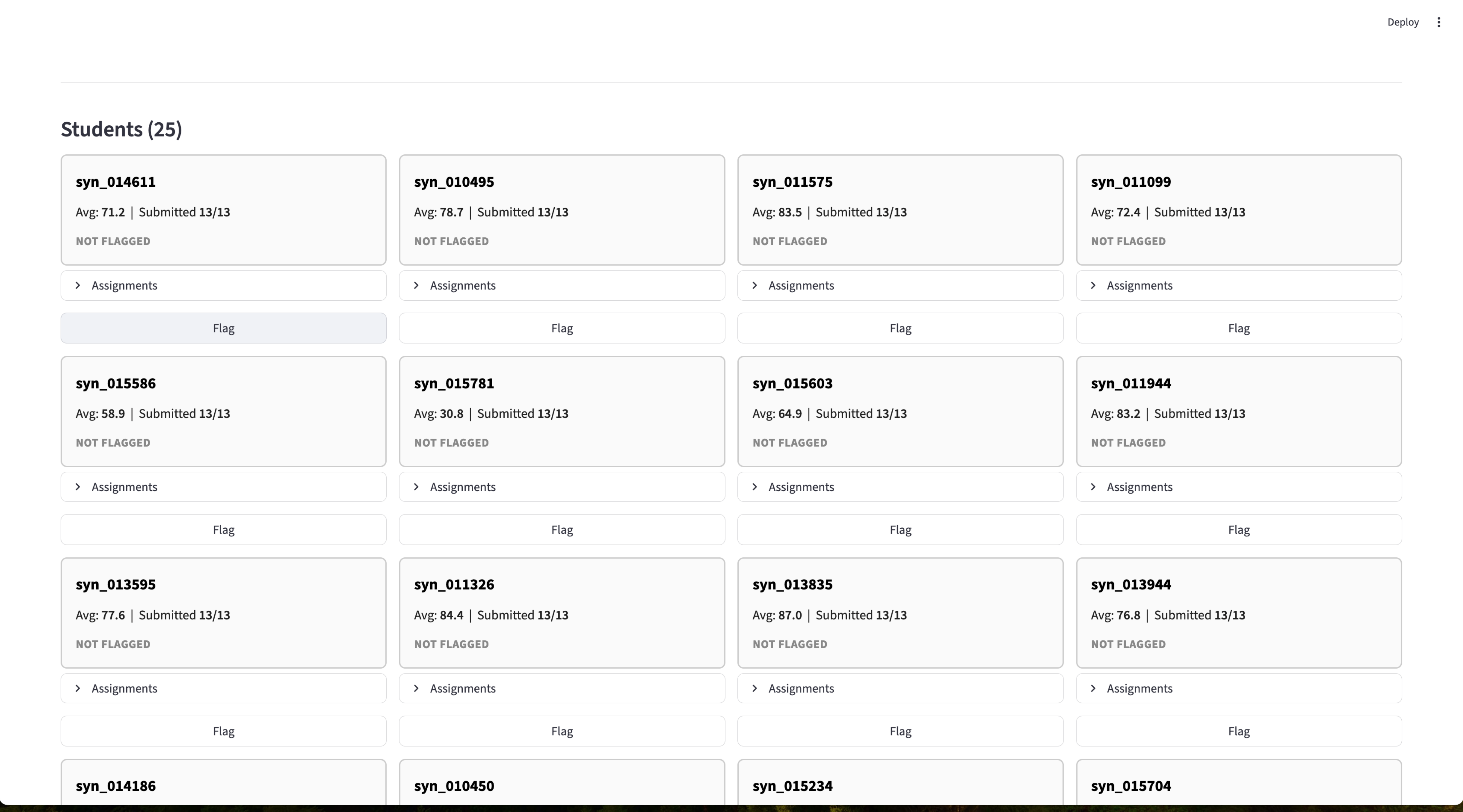}
    \caption{Page for the no-agent scenario.}
    \label{fig:human study 3}
\end{figure*}

\begin{figure*}[t]
    \centering
    \includegraphics[width=1\linewidth]{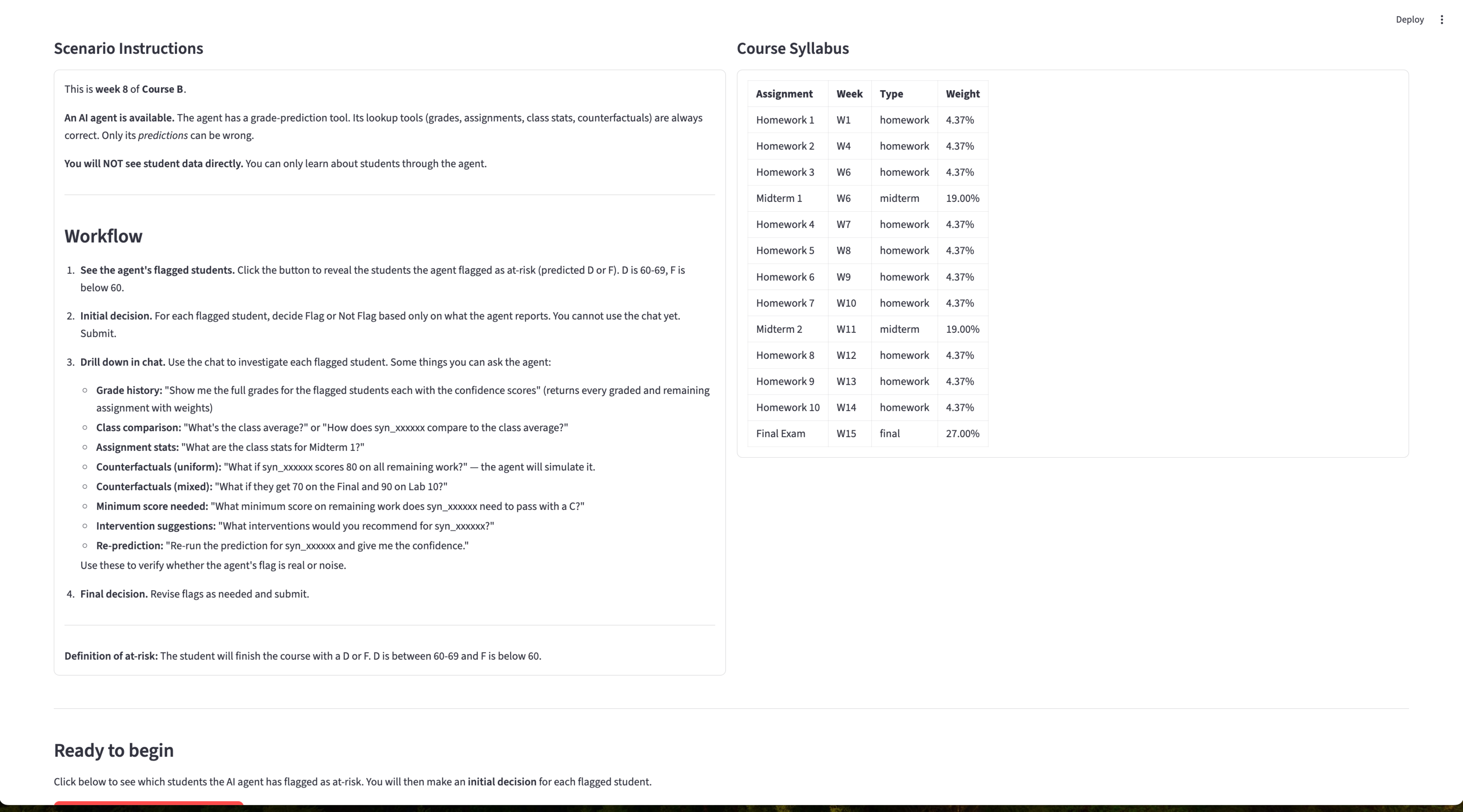}
    \caption{Instruction page for beginning the with-agent scenario.}
    \label{fig:human study 4}
\end{figure*}

\begin{figure*}[t]
    \centering
    \includegraphics[width=1\linewidth]{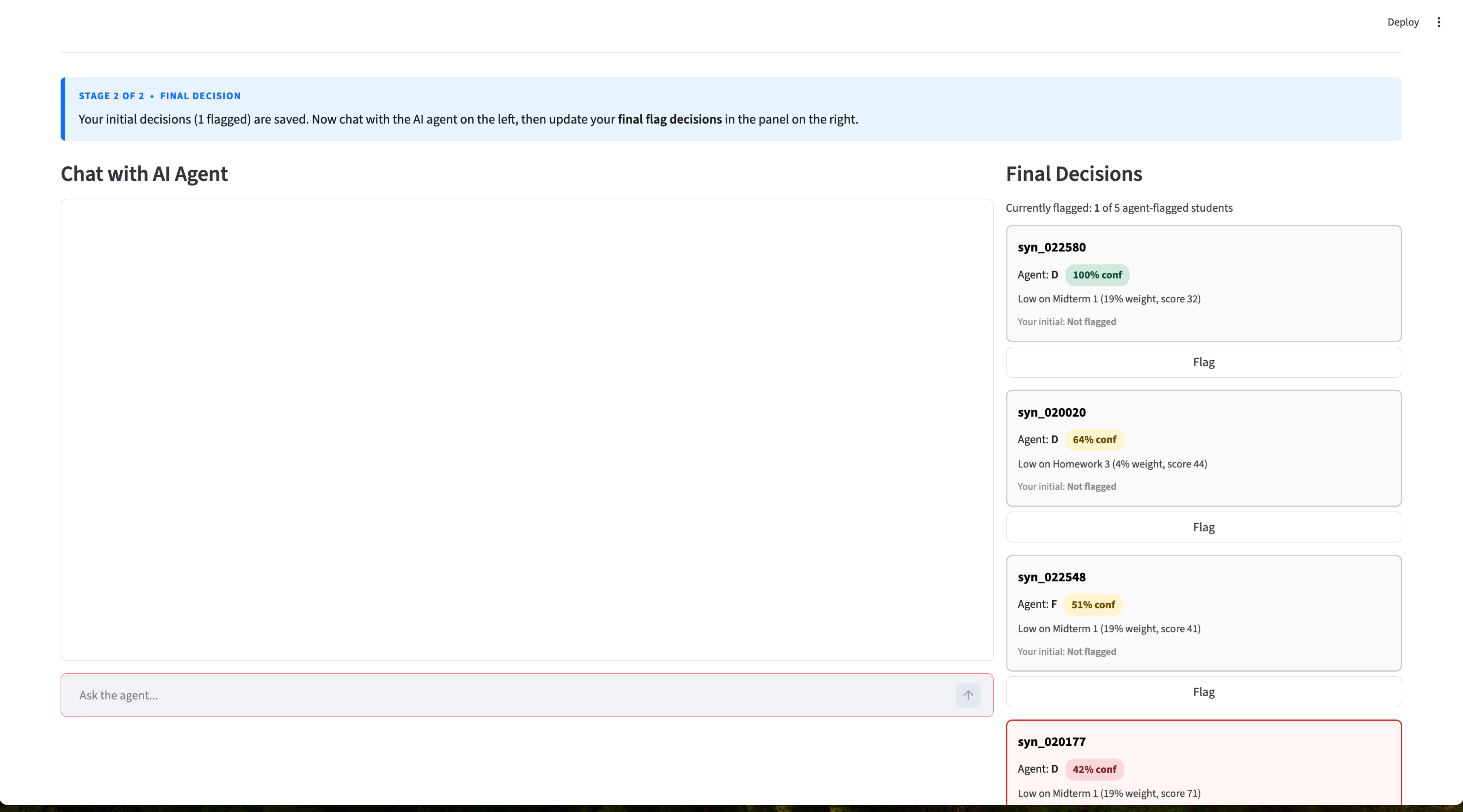}
    \caption{Page for the with-agent scenario.}
    \label{fig:human study 5}
\end{figure*}

\begin{figure*}[t]
    \centering
    \includegraphics[width=1\linewidth]{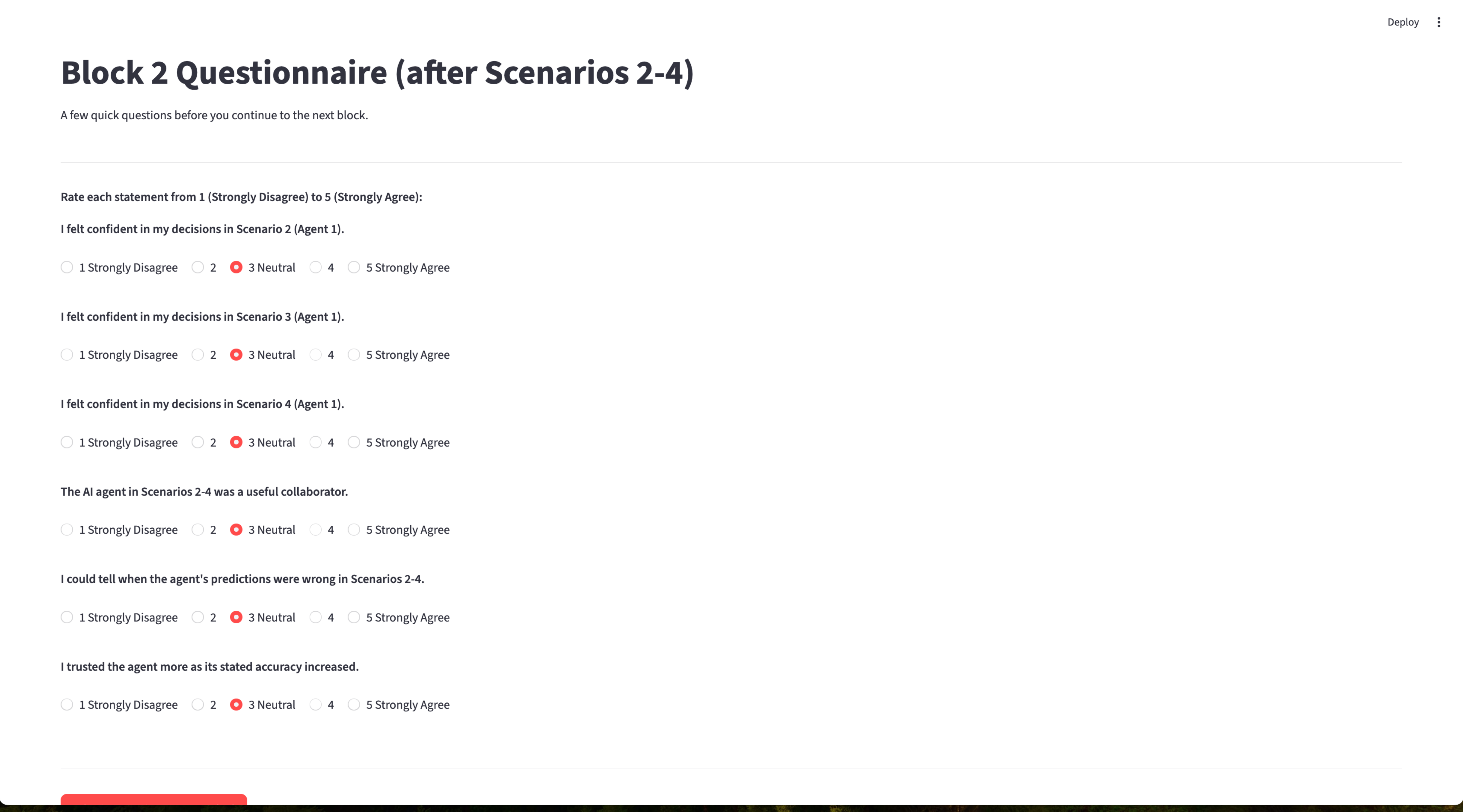}
    \caption{Questionnaire page after one with-agent scenario.}
    \label{fig:human study 6}
\end{figure*}

\begin{figure*}[t]
    \centering
    \includegraphics[width=1\linewidth]{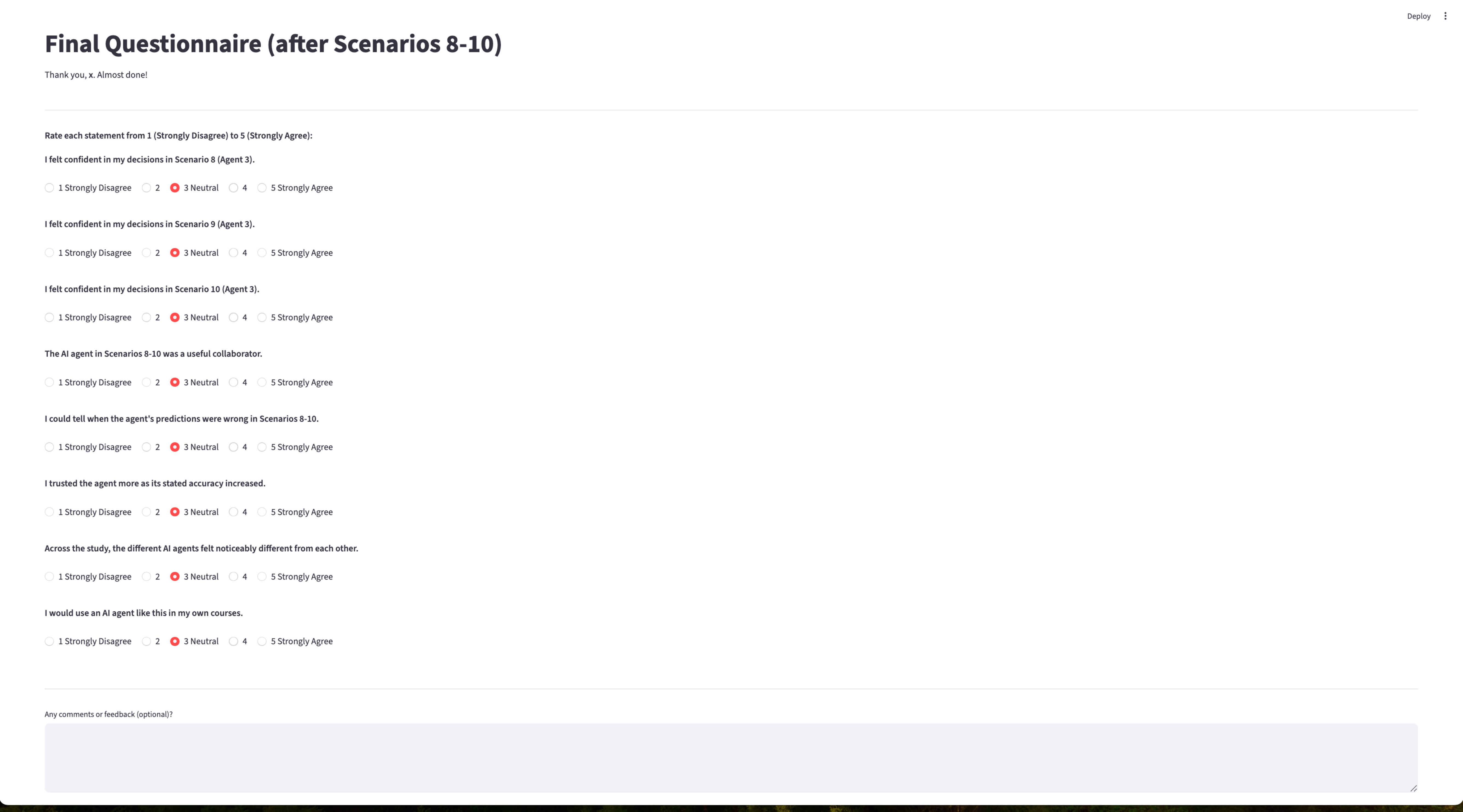}
    \caption{Questionnaire page after all with-agent scenarios.}
    \label{fig:human study 7}
\end{figure*}

\section{Human Subjects, Potential Risks, and Institutional Review}
\label{app:human-subjects-irb}

Our human study involved instructors and teaching assistants who completed
simulated academic risk prediction scenarios using the study interface described
in Appendix~\ref{app:human-study-interface}. Participants reviewed AI-flagged
students, made initial flag/no-flag decisions, interacted with an AI assistant
through a chat interface, submitted final decisions, and completed short
post-condition questionnaires. The study was conducted for research purposes
only; participant decisions did not affect any real student, course, grade, or
academic intervention. All participants were compensated.

\paragraph{Potential risks to participants.}
The study posed no more than minimal risk to participants. The main potential
risks were mild fatigue from completing multiple scenarios, possible frustration
when the AI assistant produced uncertain or incorrect predictions, and possible
concern that participants' decisions could be interpreted as an evaluation of
their personal teaching or advising ability. To mitigate these risks,
participants were informed that the scenarios were simulated, that the AI tool
could be inaccurate, and that the purpose of the study was to evaluate the
decision-support setting rather than individual participant performance.
Participants were not shown the prediction accuracy of the AI tool.

\paragraph{Risk disclosure and consent.}
Before beginning the study, participants were given study instructions describing
the task procedure, expected duration, voluntary nature of participation,
potential risks, and data use. Participants were informed that they could stop
participating at any time without penalty. They were also informed that their
responses would be analyzed in aggregate and that no personally identifying
information would be reported in the paper.

\paragraph{Privacy and confidentiality.}
We do not report personally identifying information about participants. Human
study results are reported only in aggregate, including decision quality,
reliance metrics, questionnaire responses, and interaction statistics. The paper
does not include participant names, email addresses, or individually
attributable responses. Interface screenshots included in the appendix are
representative screenshots and do not reveal participant identities.

\paragraph{Institutional review.}
Under the authors' institutional policies, this minimal-risk study using
simulated academic decision scenarios did not require full Institutional Review
Board approval. The study involved adult participants, did not affect real
students or course outcomes, and reports only aggregate, non-identifying results.

\section{License of OULAD}
\label{app: lisence OULAD}
This paper uses the data from the OULAD dataset. The OULAD dataset is released under the CC-BY 4.0 license, and we fully respect the license.



\end{document}